\documentclass{article}
\usepackage{frExamplee}
\usepackage{setspace}

\usepackage{amsmath,amsfonts,amssymb}
\usepackage{natbib}
\usepackage[binary-units=true]{siunitx}
\usepackage{mathtools}
\usepackage[pdftex]{graphicx}
\usepackage[usenames,dvipsnames]{color}
\usepackage{multirow}

\title{Synchronous-Clock Range-Angle Relative Acoustic Navigation: A Unified Approach to Multi-AUV Localization, Command, Control and Coordination}
\author{
Nicholas R.~Rypkema \\
Applied Ocean Physics \& Engineering\\
Woods Hole Oceanographic Institution\\
Woods Hole, MA 02543 USA \\
\texttt{nrypkema@whoi.edu} \\
\And
Henrik Schmidt \\
Department of Mechanical Engineering \\
Massachusetts Institute of Technology \\
Cambridge, MA 02139 USA \\
\texttt{henrik@mit.edu} \\
\And
Erin M.~Fischell \\
Applied Ocean Physics \& Engineering\\
Woods Hole Oceanographic Institution\\
Woods Hole, MA 02543 USA \\
\texttt{efischell@whoi.edu} \\
}

\begin{document}

\maketitle

\begin{abstract}

This paper presents a scalable acoustic navigation approach for the unified command, control and coordination of multiple autonomous underwater vehicles (AUVs).  Existing multi-AUV operations typically achieve coordination manually, by programming individual vehicles on the surface via radio communications, which becomes impractical with large vehicle numbers; or they require bi-directional inter-vehicle acoustic communications to achieve limited coordination when submerged, with limited scalability due to the physical properties of the acoustic channel.  Our approach utilizes a single, periodically-broadcasting beacon acting as a navigation reference for the group of AUVs, each of which carries a chip-scale atomic clock (CSAC) and fixed ultra-short baseline (USBL) array of acoustic receivers.  One-way travel-time (OWTT) from synchronized clocks and time-delays between signals received by each array element allows any number of vehicles within receive distance to determine range, angle, and thus determine their relative position to the beacon.  The operator can command different vehicle behaviors by selecting between broadcast signals from a predetermined set, while coordination between AUVs is achieved without inter-vehicle communication, by defining individual vehicle behaviors within the context of the group.  Vehicle behaviors are designed within a beacon-centric moving frame of reference, allowing the operator to control the absolute position of the AUV group by re-positioning the navigation beacon to survey the area of interest.  Multiple deployments with a fleet of three miniature, low-cost SandShark AUVs performing closed-loop acoustic navigation in real-time provide experimental results validated against a secondary long-baseline (LBL) positioning system, demonstrating the capabilities and robustness of our approach with real-world data.

\end{abstract}

\section{Introduction}

The nature of the underwater ocean environment makes it, in many ways, the ideal stage for robotic sensing and exploration -- enormous pressures, darkness, corrosiveness, biofouling, opaqueness to electromagnetic and radio waves, and its sheer size all contribute to an environment which is 
%extremely difficult for humanity to explore.
hostile to human exploration and extremely difficult to sense and measure remotely.  At the same time, various physical, chemical and biological processes that take place in the ocean over a wide range of temporal and spatial scales play a critical role in the carbon cycle \citep{Williams2011,Steinberg2017,Cermeno2008}, and accurate measurement of these processes is crucial to our understanding of the changing climate \citep{Fasham2003} and for informing numerical models \citep{Doney2004}.  The use of autonomous platforms to observe these complex ocean processes has been a dream of oceanographers since the early 1990's \citep{Curtin1993}, and the promise of such platforms for understanding the complexity of the ocean is well established; \citet{Whitt2020} presents a recent and detailed overview of the importance of in-situ ocean sampling for guiding science and policy, the issues surrounding existing observational methods, and lays out a future vision of autonomous ocean observations in which advanced robotic technologies are of vital importance.

Among these ocean processes, those that occur at the $\mathcal{O}(1$\si{\kilo\meter}$)$ submesoscale can be of particular interest to ocean scientists -- these features are known to make an important contribution to the redistribution of mass, heat and biogeochemical tracers in the upper ocean \citep{Leif2008}, with their dynamics driving local mixing and stratification that result in the exchange of properties such as nutrients and organic matter between the surface and ocean interior \citep{Pascual2017}.  Transient events like harmful algal blooms and gas hydrate plumes also occur at the submesoscale.  Whereas $\mathcal{O}(10-100$\si{\kilo\meter}$)$ mesoscale processes are large enough to be observed by remote sensing methods such as satellite altimetry \citep{LeTraon2013}, such approaches do not have the required sensitivity to resolve scales shorter than $100$\si{\kilo\meter} where in-situ methods must instead be used.  The collection of in-situ observations is a compelling use-case of autonomous underwater vehicles (AUVs), which are able to traverse submesoscale features and dynamically adapt their trajectories using sensor feedback \citep{Flexas2018}; however, these features can evolve on timescales of hours to weeks, making it difficult to attribute variations in a single-sensor stream to either changes in space or time.  Such ambiguities can be resolved through the use of multiple AUVs to synchronously sample the feature volume and capture its four-dimensional dynamics.  

While there has been significant progress in experimental multi-robot research in the past decade, this progress has largely been restricted to the ground \citep{Brambilla2013} and aerial \citep{Chung2018} domains.  The challenge for underwater robotics is that the nature of the environment presents severe limitations on navigation and communication capabilities, which restrict the use of above-water methods and algorithms for multi-vehicle coordination, command and control.  Additionally, conventional AUVs typically use a standard sensor payload for navigation, comprised of a Doppler velocity log (DVL) and a high-grade inertial measurement unit (IMU) combined into an inertial navigation system (INS) for dead-reckoning; the cost, size, and power requirements of this payload result in vehicles that are large, expensive and unwieldy to deploy.  As a consequence, there is a large gap between simulation studies and experimental research in underwater robotics, and multi-AUV deployments are exceedingly rare.  In this work, we present an approach to multi-AUV operations that unifies underwater robotic localization, command, control and coordination.  Our approach is centered around the use of a single acoustic beacon, which is used as a navigation aid and which broadcasts a simple operator command to any number of underwater vehicles within listening range.  Clock synchronization between the AUVs and the beacon, and vehicle-mounted ultra-short baseline (USBL) arrays allow each AUV to passively receive broadcasts and accurately determine their position relative to the beacon using one-way travel-time (OWTT) range and angle -- this makes our approach suitable for the emerging class of miniature, low-cost AUV, which has neither the space nor the necessary power for a DVL-aided INS and whose dead-reckoning accuracy is especially poor as a result \citep{Underwood2017, Phillips2017}.  Since our approach does not make use of bi-directional acoustic communications, it scales easily with an increasing number of AUVs.

The remainder of this paper is structured as follows: Section~\ref{sec:related_work} outlines related work on multi-AUV research with an emphasis on fielded systems, as well as prior literature on relevant underwater acoustic navigation methods.  Section~\ref{sec:acoustic_system} details the hardware and algorithmic approaches used to enable one-way travel-time inverted ultra-short baseline (OWTT-iUSBL) localization that is at the heart of the relative acoustic navigation paradigm.  Section~\ref{sec:relative_acoustic_navigation} describes how multi-AUV command, control and coordination is achieved using the single beacon OWTT-iUSBL system.  Section~\ref{sec:experiments} presents closed-loop acoustic navigation experiments and associated results from six deployments carried out over the course of three days in a river environment.  Finally, section~\ref{sec:conclusion} provides some discussion and concluding remarks.

%While individual autonomous underwater vehicles (AUVs) and gliders have successfully performed this in-situ sampling, submesoscale features can evolve within a few days, making it difficult to attribute variations in a single-sensor stream to either changes in space or time -- the use of multiple AUVs synchronously sampling over the volume of the feature would be an ideal method of capturing its complex four-dimensional dynamics.

%something something hope to provide an additional tool for observational oceanography.

\section{Related work}\label{sec:related_work}

The rapid attenuation of electromagnetic waves underwater means that the propagation length of radio-frequency communications is limited to a few tens of centimeters; as a result, many above-water approaches to multi-robot coordination that assume dense sensing for detecting neighbors, high-throughput communication to exchange information, and accurate position information for navigation \citep{Yan2013}, cannot be used for multi-vehicle coordination in the underwater domain, where acoustic sensing is often sparse and incomplete and where acoustic communication is unreliable and typically limited to $\mathcal{O}(10-100$~\si{\bit\per\second}$)$ \citep{Schneider2010}.  These limitations have led to the use of centralized, human-in-the-loop command and control approaches, where multiple AUVs are coordinated by an operator during periods when the vehicles have surfaced and radio communications are available.  The work of \citet{Fiorelli2006} provides one of the earliest demonstrations of such an approach in Monterey Bay in 2003, where three gliders were tasked to maintain formation and sample ocean temperature fronts as a group, as well as with an additional propeller-driven AUV.  With no means of inter-vehicle communication, the gliders surfaced every two hours and communicated to a central computer; centralized command and control was performed iteratively by calculating desired waypoints using a virtual potential formation control scheme at this computer, which were broadcast to the fleet on every surfacing event.  Additional results from these experiments were reported by \citet{Leonard2007}, which were used to derive further control laws for coordinated optimal sampling for a follow-up experiment in August 2006.

This system was expanded upon in the work of \citet{Paley2008} into a full centralized command and control architecture called the Glider Coordinated Control System (GCCS), in which ocean models were simulated to predict glider motion when underwater, and updated with glider global positioning system (GPS) information when they surfaced, with communications between the vehicles and the land-based GCCS occurring over the Iridium satellite network.  Predicted positions of the gliders were used to update control laws in the GCCS to perform desired missions and complete scientific objectives, with these coordinated trajectories transmitted and uploaded to the gliders upon surfacing.  The system was validated with experimental deployments in Buzzards Bay in March 2006, demonstrating successful coordination when tidal flow was weak, but performance degraded substantially during periods of strong tidal flow.

This line of work ultimately led to the largest and longest deployment of multiple underwater vehicles to date, in which \citet{Leonard2010} deployed a fleet of $10$ gliders, $6$ of which operated continuously for $24$ days, with the desire to sample intermittent upwelling events in Monterey Bay in August of 2006.  Again, the GCCS was used to predict glider trajectories using ocean models, and updated plans were transmitted to each glider upon surfacing, demonstrating sustained and automated coordinated control of these gliders.

Other examples of multi-AUV deployments include work by \citet{Das2011}, in which a propeller-driven AUV and a glider were simultaneously deployed to observe the evolution of phytoplankton blooms in Monterey Bay in 2010.  These deployments were performed as independent stages in the overall experiment, where assets were deployed each day and retrieved, their collected data analyzed in a central command room where the following mission was planned and the assets redeployed.  Another example of experimental work in which multiple AUVs and gliders were essentially programmed to operate independently via a central command was recently demonstrated by \citet{Branch2018} in May of 2017.  In these experiments, each vehicle operated independently using an adaptive behavior to detect temperature fronts; upon surfacing, data uploaded to the central command was used to update an estimate of the front location, which was then used to transmit commands to the vehicles to transect this new front estimate.  Work by \citet{Claus2018} performed in the fall of 2016 also demonstrated a similar operating paradigm in which two AUVs were operated together as independent platforms, but whose positions were augmented via range measurements to a single acoustic beacon.  The issue with approaches such as these, where vehicles are periodically coordinated via a central command and control structure upon surfacing, is that their utility is limited to vehicles that operate over large spatial and temporal length scales, such as gliders, where precise sampling is of less importance, or they require the use of vehicles with a high-grade INS to accurately maintain position.

Alternatively, field experiments have also been demonstrated in which multiple AUVs operate in a self-coordinated fashion, without the need to surface and transmit data to a centralized command.  This includes early work by \citet{Soares2013}, in which two leader AUVs transmitted their heading information via acoustic modems to a follower AUV; these bi-directional acoustic messages enabled the follower to determine its range to both leaders and maintain position between them, as demonstrated in a saltwater bay in Lisbon in June of 2012.  Early work by \citet{Petillo2014} also experimentally demonstrated the use of two AUVs for the adaptive detection and characterization of internal waves in the Tyrrhenian Sea in August of 2010; during these experiments, a follower AUV adaptively trailed a leading AUV, with both vehicles dynamically modifying their depths to maintain position within a thermocline in order to characterize the internal wave structure.  Acoustic modems were used to communicate relevant information between the two vehicles as well as a topside ship-based command center.  Other experiments include those performed by \citet{Walls2015}, in which two AUVs demonstrated a cooperative localization scheme where OWTT range factors between the vehicles were communicated to each other acoustically, and used within a factor graph framework for localization.  More recently, in 2013 and 2014, \citet{Lin2017} performed a two-AUV deployment, where each vehicle was equipped with a stereo hydrophone system to estimate the relative distance and bearing to an acoustic transmitter tag to track marine life; the AUVs exchanged their estimate of the tag location via acoustic modems, allowing each to incorporate both estimates into a joint solution. The issue with these in-situ self-coordinated approaches is that they require active two-way communications between vehicles, with limited scalability for large groups of AUVs due to the limited bandwidth and unreliable nature of the acoustic channel.

An additional challenge for underwater robotics is the fact that rapid electromagnetic absorption and the unstructured environment make localization a particularly difficult task \citep{Paull2014}.  As GPS is not available underwater, AUVs typically navigate using dead-reckoning with velocity measurements from a DVL and attitude measurements from an IMU; however, navigating in this manner is always subject to an unbounded increase in positioning error, with error ranging from anywhere between $0.05\%$ to $20\%$ of distance traveled depending on the quality of the sensors used, the absence of a DVL or if the DVL has bottom-lock, and environmental conditions.  To bound this error, AUVs either periodically surface for a GPS fix, or make use of an external position reference, which is usually an acoustic positioning system such as a static long-baseline (LBL) network of beacons or a ship-mounted USBL system.  One major drawback of traditional LBL and USBL systems is that they require query-response two-way travel-time (TWTT) ranging, which increases vehicle power-use and takes up acoustic bandwidth that would be better used for communication.  Although work has been done in integrating localization data into acoustic communication packets to reduce channel use, both in the context of moving LBL \citep{Munafo2017} and USBL systems \citep{Costanzi2017}, TWTT systems are still channel-limited in the number of vehicles that can be supported.

Recent work has focused on the use of synchronized clocks in order to overcome this issue of scalability -- instead of a query-response architecture, this enables one-way travel-time (OWTT) ranging through precise knowledge of the transmit and receive times of the acoustic packet.
Initial work by \citet{Eustice2006} successfully demonstrated that OWTT range measurements from a single beacon could aid AUV navigation; in these experiments, recorded OWTT ranges were combined with beacon position in an offline, least-squares batch optimization to estimate global AUV position.  This work was improved upon by \citet{Webster2012}, showing offline post-processing of single-beacon OWTT range measurements using a centralized extended Kalman filter (EKF), which could be used to bound dead-reckoning drift on-board an AUV over kilometer spatial scales.  Closed-loop AUV navigation using this single-beacon OWTT ranging approach was recently validated through experiments by \citet{Claus2018}, in which the beacon broadcast acoustic packets containing its own position information, and this data was fused with range information by a particle filter and EKF on-board the vehicle to perform global self-localization.

Single-beacon OWTT ranging approaches are limited in the fact that they require multiple range measurements from a variety of relative bearings in order to attain a suitably unambiguous positional fix.  Alternatively, range measurements from two or more beacons can be used as a OWTT LBL positioning system, which provide a fully-determined position on every broadcast via multilateration; such an approach has been demonstrated by \citet{Melo2016} as well as more recently by \cite{Quraishi2019} and also \citet{Randeni2020}.  A variation of this approach using a OWTT, moving LBL system was very recently demonstrated by \citet{Simetti2021} to support navigation for a fleet of four AUVs, each towing an $8$\si{\meter} long hydrophone streamer for seismic acquisition.  In this work, each AUV was equipped with two acoustic modems -- an $18-34$~\si{\kilo\hertz} mid-frequency modem that received GPS position information and OWTT ranges from corresponding modems mounted on two autonomous surface vehicles (ASVs) acting as anchors for moving OWTT LBL positioning; and a $42-65$~\si{\kilo\hertz} high-frequency modem for communication of data, AUV monitoring, and reception of operator commands.  This complete system demonstrated integrated localization, command and control of all four AUVs, and along with highly accurate clock synchronization via chip-scale atomic clocks (CSACs), enabled fully automated acquisition of seismic images.  The particular application of seismic acquisition demonstrated by this work illustrates the complementary advantage of synchronized clocks for both underwater navigation and environmental observation -- time-synchronization is necessary to temporally align observations across a fleet of vehicles, and also enables multi-vehicle localization in a scalable manner.

The moving OWTT LBL approach demonstrated by \citet{Simetti2021} carries with it an associated cost in algorithmic and operational complexity from the use of multiple beacons -- the global position of each beacon must be broadcast, then interpreted and integrated into the localization estimate of individual AUVs.  An alternate approach, first proposed by \citet{Jakuba2015}, is to determine OWTT range and angle to a single acoustic beacon, and is known as one-way travel-time inverted ultra-short baseline (OWTT-iUSBL) localization.  As with OWTT LBL, this approach provides an unambiguous position estimate on every transmission -- with each AUV equipped with a USBL array, range and angle to the beacon can be calculated allowing each vehicle to determine its relative position from the beacon.  We have previously reported successful experiments using this approach for both absolute positioning with a static beacon \citep{Rypkema2017}, as well as relative positioning with a moving beacon \citep{Rypkema2018}, demonstrating its utility as a navigational method for low-cost, miniature AUVs.  Subsequent work \citep{Fischell2019} reported initial findings in using OWTT-iUSBL for command and control of multiple AUVs, where each vehicle operated in a beacon-centric frame of reference, allowing the operator to control the location of the fleet through movement of the beacon.  While previous papers have reported on aspects of our OWTT-iUSBL system, this paper contributes significant additions: First, a detailed description of the hardware and algorithms for acoustic processing and navigational filtering are provided for the acoustic navigation system.  Second, we detail the approach used to command, control and coordinate multiple AUVs using the OWTT-iUSBL system.  Finally, we provide extensive experimental results showing the system in use over six separate deployments, with closed-loop navigation results validated against a secondary LBL system.

\section{One-way travel-time inverted ultra-short baseline (OWTT-iUSBL) acoustic positioning}\label{sec:acoustic_system}

\begin{figure}[!t]
    \centering
    \includegraphics[width=.6\textwidth]{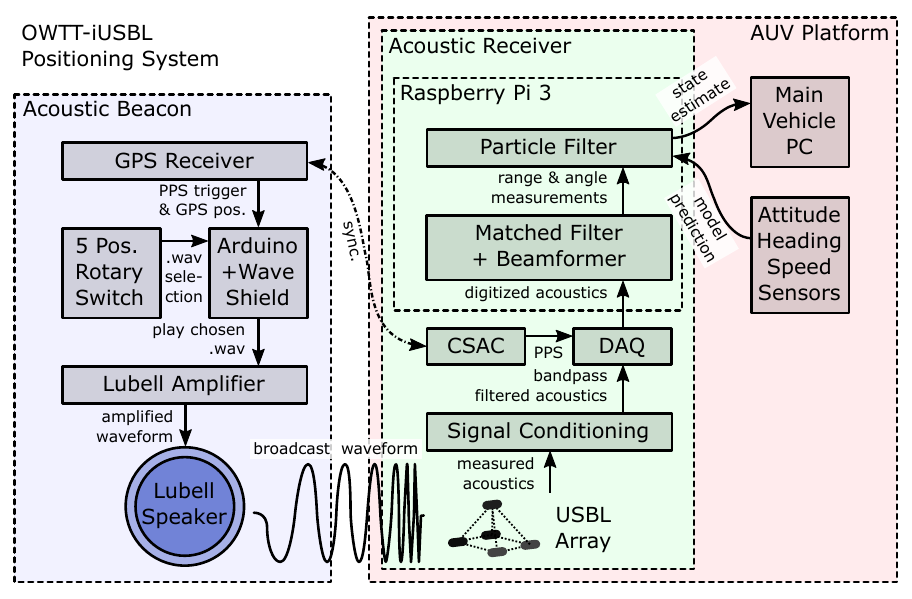}
    \caption{OWTT-iUSBL system diagram, illustrating relevant hardware and software processing components.}
    \label{fig:system_diagram}
\end{figure}

Coordination, command and control of multiple AUVs in our approach is made possible by a custom-made one-way travel-time inverted ultra-short baseline (OWTT-iUSBL) acoustic navigation system, the details of which are described in this section.  Figure~\ref{fig:system_diagram} illustrates the main hardware and software components of the OWTT-iUSBL system, consisting of a single acoustic beacon, the USBL acoustic receiver, and the underwater robotic platform.  In short, the acoustic beacon broadcasts a user-defined signal at the start of every second, triggered by the pulse-per-second (PPS) signal from a GPS receiver;  precise time-synchronization with a chip-scale atomic clock (CSAC) allows the acoustic receiver to synchronously record and digitize the broadcast signal on all elements of its USBL array every second; finally, these digitized signals are processed using matched filtering and beamforming to obtain range and angle measurements, which are fused with platform speed and attitude measurements using a closely-coupled particle filter to generate a temporally-consistent state estimate for navigation.

\subsection{Hardware}

While the software and algorithms described in this paper are largely agnostic to the particular hardware used, the hardware components of our OWTT-iUSBL implementation are detailed here in the interest of allowing the reader to replicate the system.  As the majority of the components are commercial off-the-shelf, this described system can also be used as a template for a generic acoustic receiver or positioning system.

\subsubsection{Beacon}

The OWTT-iUSBL custom acoustic beacon is comprised of the components shown to the left of figure~\ref{fig:system_hardware}.  It uses the rising edge of the PPS signal from a GPS receiver to trigger the playback of one-of-four user-defined WAV files stored on an Adafruit Wave Shield audio board attached to an Arduino microcontroller (MCU); the output of this audio board is connected to a Lubell 3400 $60$ \si{\watt} amplifier, which amplifies the acoustic signal and broadcasts it into the water via a Lubell LL916C underwater speaker.  Beacon electronics and the amplifier are housed in a Pelican 1120 Protector Case for water-resistance, with the speaker cable routed through a hole from the amplifier to the speaker, allowing the user to lower the speaker into the water over the side of a motorboat or from a dock.  A 5-position rotary switch mounted on the side of the Pelican case allows the user to manually select between the four stored waveforms, as well as no transmission.  The result is an underwater acoustic beacon that can periodically broadcast one-of-four user-defined waveforms at a rate of $1$\si{\hertz} with a jitter of less than $1$~\si{\milli\second}, and whose position can be logged by GPS.  Although jitter can be improved through custom designed circuitry, the Arduino MCU and Wave Shield provide an accessible and easily-replicable electronics design for a highly-capable beacon that can transmit any user-designed signal between $0.2-23$~\si{\kilo\hertz}.

\subsubsection{Ultra-short baseline receivers}

\begin{figure}[t!]
    \centering
    \includegraphics[width=.95\textwidth]{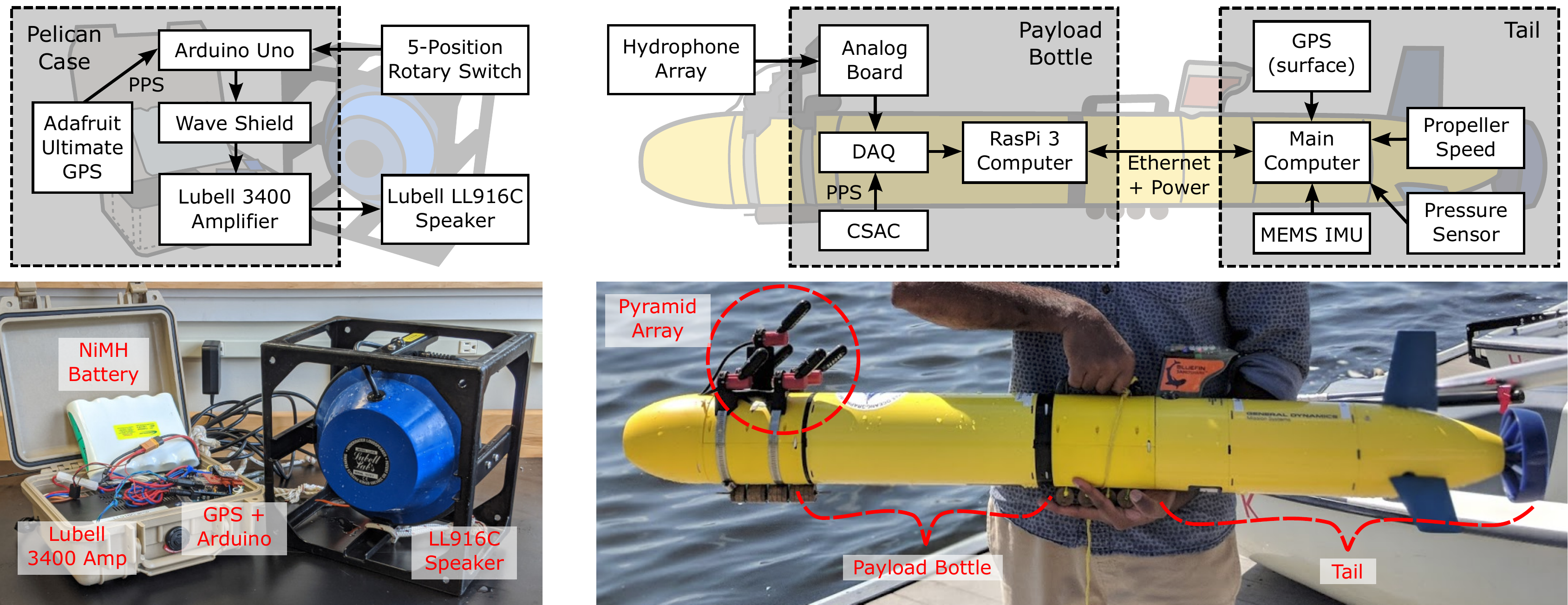}
    \caption{OWTT-iUSBL beacon hardware (left) and acoustic receiver and SandShark AUV platform (right).}
    \label{fig:system_hardware}
\end{figure}

OWTT-iUSBL receivers have five main hardware components, shown to the right of figure~\ref{fig:system_hardware}.  Each receiver includes a USBL array, consisting of five High Tech Inc. HTI-96-Min hydrophones with current-mode pre-amplifiers.  This array is rigidly attached above the nose of the vehicle in a pyramidal arrangement of $8$~\si{\centi\meter} edge length.  The acoustic energy the array measures is converted into a voltage signal and bandpass-filtered using a passive resistor-capacitor circuit ($0.01 \leq f_{bp} \leq 160 $~\si{\kilo\hertz}), before being converted into a digital signal using a Measurement Computing USB-1608FS-Plus digital acquisition device (DAQ).  In order to synchronously start the digital conversion in sync with the broadcasts of the beacon, the DAQ is triggered to record using the rising edge of a PPS signal generated by a Microsemi SA.45s CSAC that is synchronized to GPS PPS prior to vehicle launch.  These CSACs will typically drift by less than $100$~\si{\micro\second} per $24$~\si{\hour} in holdover, assuring accurate synchronization while the AUV is underwater, with a jitter of approximately $80$~\si{\pico\second} \citep{Kim2017}.  The DAQ is programmed to record a user-defined number of samples of acoustic data at a user-specified sampling rate from each element of the array after each triggering event, with this raw acoustic data made available to an on-board Raspberry Pi 3 computer.  In this work, the DAQ was configured to record $8000$ samples per element at a sampling rate of $37.5$~\si{\kilo}S\si{\per\second}, translating to a maximum sensing range of approximately $316$~\si{\meter}.  When combined with vehicle attitude measurements from compass and inertial sensors on the AUV, acoustic range and angle measurements in the body-fixed frame can be used to estimate the relative $(x,y)$ position of the beacon in the vehicle-carried East-North-Up local-level frame; if the position of the beacon is known in the absolute frame, the vehicle can subsequently be localized in the absolute frame of reference.  The electronics of the acoustic receiver are housed in a $200$~\si{\meter} depth-rated dry payload bottle that make up the front half of the vehicle, with SubConn Micro Circular cables providing connections to the leads of the USBL array as well as a connection to the rear half of the AUV to receive power and Ethernet communications.

\subsubsection{Autonomous underwater vehicles}

The underwater robotic platform used in this work is the $200$~\si{\meter} depth-rated, low-cost, miniature SandShark AUV from Bluefin Robotics \citep{Underwood2017}.  The rear half of the vehicle, as shown in figure~\ref{fig:system_hardware}, is the standard tail section provided by Bluefin, and is equipped with attitude and heading sensors, a thruster, and control surfaces.  The AUV is propelled using a single magnetically-coupled thruster, which provides a measure of vehicle speed via an empirically-derived mapping of thruster rotations-per-minute (RPM) to speed-over-ground.  Three stepper motors actuate three control fins laid out in a triangle configuration to enable active roll control as well as pitch and heading control.  The vehicle mast contains GPS and WiFi receivers for global positioning and communications when the vehicle is surfaced, and LED lights for status indication.  Sensors include a Sparton 9--axis microelectromechanical (MEMS) IMU with magnetometer for attitude and heading estimation, a pressure sensor to measure vehicle depth, and an Imagenex Model 852 echo sounder to estimate vehicle altitude.  A Linux-based main vehicle computer serves as the brains of the vehicle, and by default uses attitude, heading and speed measurements to integrate vehicle position over time via dead-reckoning to navigate when not on the surface.  The pitch ($\beta$) compensated RPM-to-speed-over-ground conversion is given by the simple scaled conversion:

\begin{align}
\mathsf{sog}_{v} &= \mathsf{RPM}\cdot 1.25 \times 10^{-3} \cdot \cos(\beta).
\label{eq:speed_over_ground}
\end{align}
Note that as the AUV is not equipped with a DVL, dead-reckoned position error grows quickly, at a rate of approximately $2\%-10\%$ of distance traveled (or approximately $1-6$~\si{\meter\per\minute} at a speed of $1$~\si{\meter\per\second}) in environments with low external disturbances.  Experimental comparison with a Hemisphere V102 dual-antenna GPS indicated that one standard deviation of heading error for these vehicles is approximately $3$\si{\degree} after magnetometer and IMU calibration.  A lithium-ion battery in the tail-section powers all vehicle systems, including the payload via an underwater SubConn cable that connects the tail and payload sections, and which allows communication and transfer of sensor data between the main vehicle computer and the payload Raspberry Pi 3.

To enable payload control of the AUV, we make use of the frontseat--backseat paradigm, which cedes all low-level vehicle control to the main vehicle computer in the tail section.  The main vehicle computer provides frontseat state information and sensor data to the Raspberry Pi 3 payload backseat computer via the Bluefin Standard Payload Interface \citep{Goldberg2011}.  As illustrated in figure~\ref{fig:system_diagram}, the backseat computer is tasked with performing the necessary acoustic processing and Bayesian filtering for OWTT-iUSBL navigation, providing this navigation solution to the MOOS-IvP autonomy framework \citep{Benjamin2010} for computation of desired speed, heading and depth setpoints, and transmission of these setpoints to the frontseat computer for execution of low-level AUV control.  When equipped with the acoustic receiver payload, the AUV diameter is $12.4$~\si{\centi\meter}, with an overall length of around $105$~\si{\centi\meter} and an in-air weight of approximately $15$~\si{\kilo\gram}.

\subsection{Acoustic signal processing}

The acoustic data recorded and digitized by the OWTT-iUSBL receiver can be processed to generate an estimate of the relative position of the beacon in the reference frame of the USBL array.  This processing consists of two procedures: the use of a matched filter to detect the presence and onset time of the broadcast signal from the start of the second for range estimation; and beamforming to phase-align the signals recorded by each array element for angle estimation. 

\subsubsection{Range measurement}

\begin{figure}[!t] 
	\centering
	\includegraphics[width=0.9\textwidth]{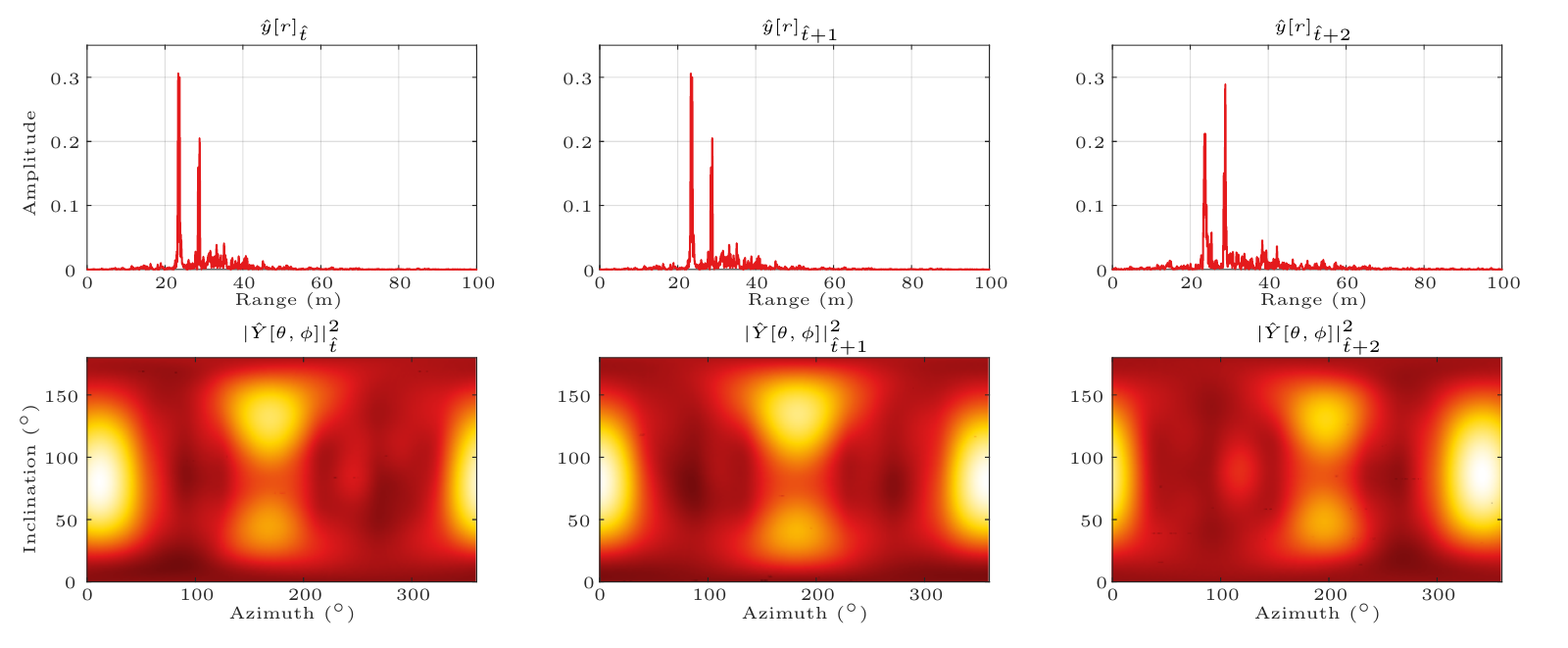}
	\caption{Three-second sequence of measurement pseudo-distributions from acoustic processing for range (top) and angle (bottom) using the SandShark AUV acoustic receiver.} 
\label{fig:acoustic_measurement_distributions}
\end{figure}

Range measurements are generated as the combined outputs of applying the matched filter to the signal measured on each array element.  The matched filter is known to be the optimal linear filter for performing signal detection in the presence of additive white Gaussian noise through signal-to-noise ratio (SNR) maximization \citep{Wainstein1962}, and is in essence the process of correlating the signal against a template of the waveform to be detected.

The acoustic data for each element ($x_i[n]$) is first pre-whitened and magnitude-normalized using the phase transform (PHAT), as all relevant information exists within signal phase:

\begin{equation}
\hat{X_i}[\omega] = \frac{X_i[\omega]}{|X_i[\omega]|} \label{eq:II_phat}
\end{equation}
where $X_i[\omega]$ is the Fourier transformed signal received by element $i$.  The PHAT has been empirically shown to improve robustness to noise and reverberation in real-world environments \citep{Knapp1976}.  The matched filter output is then given by:

\begin{align}
y_i[n] &= \sum_{k=-\infty}^{\infty} \hat{x}_i[k] s[k - n] = \hat{x}_i[n]*s[-n] \quad \Leftrightarrow \quad Y_i[\omega] = \hat{X_i}[\omega] S^{*}[\omega] \label{eq:II_discretematchedfilterpractical}
\end{align}
where $s[n]$ is a template of the broadcast signal to be detected within the acoustic data, and $S[\omega]$ is its Fourier transform.  To exclude range outliers that are caused by the loss of signal on a subset of elements due to occlusion by the AUV body, a consistency metric is enforced to retain only valid acoustic data; data is retained only if the arg-maxima of matched filter outputs from all five array elements fall within $15$ samples of each other (this is equivalent to $0.4$~\si{\milli\second} at our sampling rate of $37.5$~\si{\kilo}S\si{\per\second}).  If the measured acoustic data is deemed valid, the data from all elements is combined:

\begin{align}
\hat{y}[n] &= \sum_{i=1,j=i+1}^5 \left| y_i[n] \right| \left| y_j[n] \right| \qquad i \neq j .  \label{eq:II_combinedmatchedfilter}
\end{align}
Finally, to obtain a pseudo probability distribution for range, this combined output is normalized to unit energy, and this pseudo-distribution is transformed into the range domain by converting sample numbers ($n$) into ranges ($r$):

\begin{align}
\hat{y}[r] &= \frac{\hat{y}[r]}{\sqrt{\sum_{\infty}^{\infty} |\hat{y}[r]|^2 }} \quad \textsc{where} \quad r = \frac{c}{F_s} \cdot n = \frac{1481 \enspace \text{m/s}}{37500 \enspace \text{S/s}} \cdot n . \label{eq:II_normalizematchedfilter}
\end{align}
Example range measurement pseudo-distributions over three sequential seconds are shown in the top plots of figure~\ref{fig:acoustic_measurement_distributions}.  These plots illustrate how undesirable acoustic effects such as reflection and reverberation result in multiple peaks in the range measurement, demonstrating the need for suitable temporal filtering in order to discard false maxima.

\subsubsection{Angle measurement} \label{sec:angle_measurement}

To estimate the angle between the acoustic receiver and the beacon, an angular measurement distribution must be generated that reflects the most likely direction of the incoming acoustic signal, with its maximum ideally found in the true direction of the beacon.  A number of approaches exist to perform this direction-of-arrival (DOA) estimation, including classical and adaptive beamforming methods, subspace-based methods, and time-difference-of-arrival (TDOA) methods.  In this work, we have the advantage of assuming that only a single source is transmitting a tracking signal within the frequency band of interest -- consequently, estimation techniques that are well capable of distinguishing multiple sources, such as adaptive beamforming and subspace-based methods, are unnecessarily complex and computationally demanding for our application; and while TDOA approaches are computationally efficient, they require fast sampling rates for high angular resolution and are very sensitive to timing errors, resulting in DOA solutions that are generally less robust than beamforming approaches to acoustic noise and reflections commonly encountered in underwater environments.

In this work, we employ the approximate beamforming approach recently presented by \citet{Rypkema2020} to perform DOA estimation, known as sensor pair decomposition beamforming (SPD-BF).  This algorithm was specifically developed to localize strong acoustic sources using memory and computationally limited embedded systems, and operates by essentially combining the outputs of conventional beamforming (CBF) performed on each pair of array elements.  The general idea behind beamforming is to find the phase shifts (or time delays) necessary in order to best align the signals measured by each array element -- with the proper phase-alignment, the signals add constructively and the beamformer output power is maximized \citep{Trees2002}; the phase-shifts that produce this condition correspond to the most likely direction to the acoustic source.  The time delays $\tau_i$ of a plane wave incident onto an arbitrary array with element positions $\boldsymbol{p}_i$ from direction $\boldsymbol{a}$ with velocity $c$ are given by:

\begin{align}
\tau_i = \frac{\boldsymbol{a}^T \boldsymbol{p}_i}{c}
\quad \textsc{where} \quad
\boldsymbol{a} &= \left(\begin{array}{cc} 
-\sin(\theta) \cdot \cos(\phi)\\
-\sin(\theta) \cdot \sin(\phi)\\
-\cos(\theta)
\end{array}\right) .
\label{eq:timedelay}
\end{align}
In CBF, the beamformer is steered in different directions with the goal of finding the look-direction that will undo these time delays.  The output of a narrowband beamformer with wave frequency $\omega$ and steered in a given look-direction $(\theta,\phi)$ for an $N$-element array is computed as:

\begin{align}
Y[\omega;&\theta,\phi] = \sum_{i=1}^N{H_i[\omega;\theta,\phi] \cdot X_i[\omega]} \nonumber \\
&\textsc{where} \quad H_i[\omega;\theta,\phi] = e^{j\omega\tau_i} .
\label{eq:phaseshifts}
\end{align}
where $X_i[\omega]$ is the Fourier transformed signal received by element $i$.  In this work, we utilize wideband beamforming, and sum the power of these outputs over a range of $M$ frequencies:

\begin{align}
|\tilde{Y}[\theta,\phi]|^2 &= \frac{1}{M}\sum_{k=1}^M{|Y[\omega_k;\theta,\phi]|^2} .
\label{eq:cbfoutputpre}
\end{align}
To generate a pseudo-distribution over angle, this output is calculated for a set of look directions, producing an angle measurement whose maximum ideally occurs at the true direction to the beacon.  A significant disadvantage of CBF is that the computation time is proportional to the number of look directions over which the beamformer is steered; additionally, to speed up computation time, the phase shifts given by $H_i$ in equation~\ref{eq:phaseshifts} are typically pre-computed and stored over all look directions and $M$ frequencies, resulting in a memory usage that scales with both number of look directions and bandwidth.  

\begin{figure}[!t] 
	\centering
	\includegraphics[width=1\textwidth]{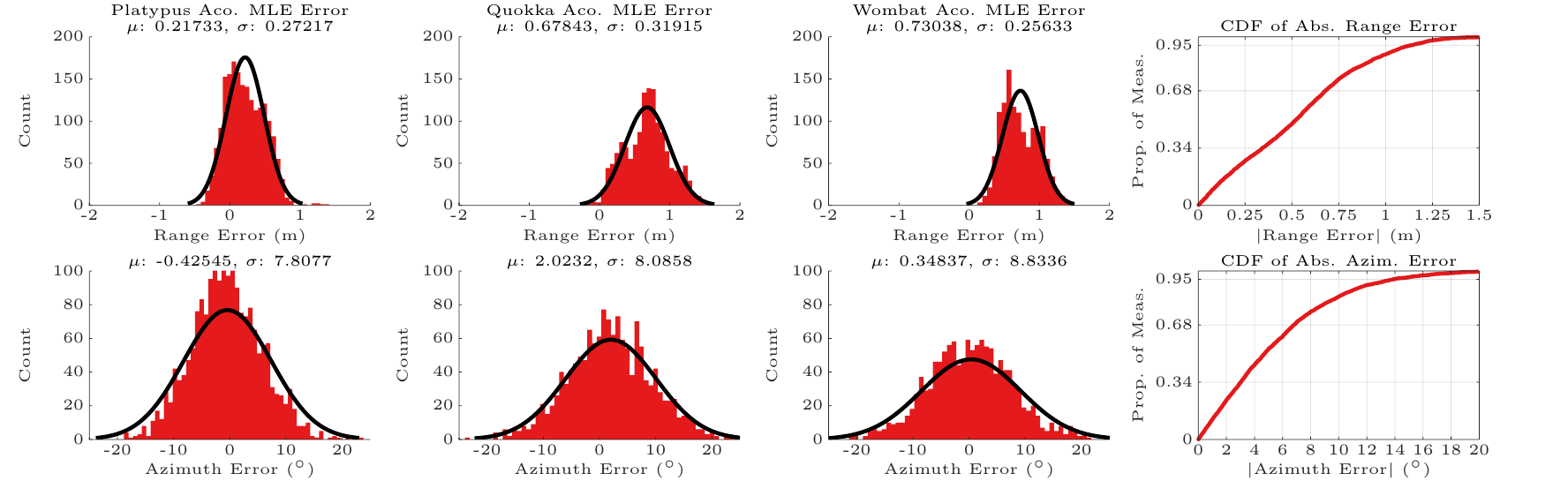}
	\caption{Statistics for OWTT-iUSBL acoustic receiver range and azimuth measurements, with histograms of range and azimuth MLE values on the left for individual SandShark AUVs, and empirical cumulative distribution functions (CDFs) for absolute range and azimuth errors on the right.} 
\label{fig:range_azim_hist}
\end{figure}

This limitation poses a problem, since the generation of an accurate and precise angle measurement requires that the beamformer be steered in very small increments over azimuth ($\phi$) and inclination ($\theta$).  The problem is that a linear increase in the resolution of $\phi$ and $\theta$ results in a quadratic increase in the number of $(\theta,\phi)$ combinations -- thus, given the $1$\si{\giga\byte} memory limit of the payload Raspberry Pi 3 computer, the use of conventional beamforming severely restricts the attainable angular resolution of the angle measurement and usable frequency bandwidth of the OWTT-iUSBL system.  Sensor pair decomposition beamforming (SPD-BF) overcomes this limitation through the insight that 1D linear arrays are steered over a single dimension parameterized by the conical angle ($\zeta$), rather than over the two dimensions of azimuth and inclination $(\theta,\phi)$; equation~\ref{eq:timedelay} thus reduces to:

\begin{align}
\tau_i = \frac{-\cos(\zeta) z_i}{c}
\end{align}
where $z_i$ is the 1D position of array element $i$.  SPD-BF decomposes the 3D array into all unique element pairs, and performs conventional beamforming over a set of conical angles for each pair; these CBF outputs are then summed over all element pairs after a nearest-neighbor transformation of the conical angles to the desired set of azimuth and inclination directions.  SPD-BF produces a result that is an approximation of CBF, but with the advantage of an enormous reduction in memory and computational cost for arrays with few elements.  Instead of quadratic growth as in the case of CBF, the computational and memory cost of SPD-BF grows in proportion to the number of conical angles used and the number of unique element pair combinations of the array.  Consequently, it enables high-resolution DOA estimation over wide bandwidths -- in this work, we perform SPD beamforming at a conical angle resolution of $0.25^\circ$ and a frequency resolution of approximately $2$~\si{\hertz}, using various broadcast signals of $2$~\si{\kilo\hertz} bandwidth in the range of $7-11$~\si{\kilo\hertz}.  The reader is referred to \citep{Rypkema2020} for greater detail on the SPD-BF method as well as its advantages and disadvantages.  Example angle measurement pseudo-distributions resulting from the use of this algorithm with our pyramidal array are shown at the bottom of figure~\ref{fig:acoustic_measurement_distributions}, illustrating the multimodal nature of these measurements.

Calibration of the OWTT-iUSBL acoustic receiver payloads on our SandShark vehicles was performed through the use of an 80/20 extruded aluminum rotational rig, in which the AUV was clamped to the submerged end of a pole at a depth of $2$~\si{\meter}, while a Hemisphere V102 dual-antenna GPS (DGPS) was rigidly attached to the above-water end of the pole to capture ground-truth azimuth; this rig allowed the vehicle to be manually rotated around $360^\circ$ in heading at a known distance from the acoustic beacon, providing a dataset to compare acoustic azimuth and range maximum likelihood estimates (MLEs) to DGPS ground-truth -- this calibration process was performed at two different ranges for all three vehicles: once at approximately $30$~\si{\meter} between the beacon and AUVs, and a second time at approximately $57$~\si{\meter}. Statistics for these acoustic range and azimuth measurements are plotted in figure~\ref{fig:range_azim_hist}, for the three individual SandShark AUVs used in this work, named \emph{Platypus}, \emph{Quokka} and \emph{Wombat}, as well as cumulative distribution functions (CDFs) of all AUV statistics combined.  These figures indicate that $68\%$ of acoustic range MLEs have an absolute error of less than $0.7$~\si{\meter}, and $68\%$ of acoustic azimuth MLEs fall below an absolute error of $7.0^\circ$.

It should be noted that the acoustic processing pipeline makes use of a couple of simplifying assumptions that, while arguably valid for the operating environment used in this work, may not be valid for other environments.  In this work, the assumption of a constant soundspeed is reasonable, since the range between the beacon and the AUVs generally remains under $100$~\si{\meter} and the water column in the river environment within which our experiments take place is homogeneous -- as such, any ray-bending that occurs between the beacon and receivers is likely to be a minor source of error in our range estimates as compared to other acoustic factors such as multi-path; the use of an estimated effective soundspeed in our work was also necessary, since our fleet of AUVs lack conductivity-temperature-depth (CTD) sensors to directly measure water density and soundspeed.  In addition, the range and azimuth measurement statistics detailed in figure~\ref{fig:range_azim_hist} are only valid for the short ranges used in our experiments, where issues such as ray-bending and sound absorption can be effectively ignored -- the SNR of received signals remain high throughout our experiments regardless of range.  These assumptions may not be valid for other missions, such as deep-water deployments or situations where large ranges are expected between the acoustic beacon and receivers.  At multi-\si{\kilo\meter}-scale depths and ranges, large changes in soundspeed over the water column can cause significant ray-bending -- consequently, the travel-time of the acoustic signal, and thus the estimated range, are not directly proportional to the Euclidean distance between beacon and receiver; the resulting range errors can be mitigated by directly measuring and compensating for the soundspeed profile \citep{Webster2012}.  At large ranges, the variation in soundspeed can also cause the acoustic signal to be concentrated and to be lost in certain regions, leading to convergence zones and shadow zones \citep{Jensen2011}; large transmission loss within shadow zones greatly reduces the acoustic SNR, degrading signal processing performance -- such range-dependent environmental effects must be seriously considered when applying the approach described here to deep-water, long-range deployments.

\subsection{Particle filter}

\begin{figure}[!t]
    \centering
    \includegraphics[width=.8\textwidth]{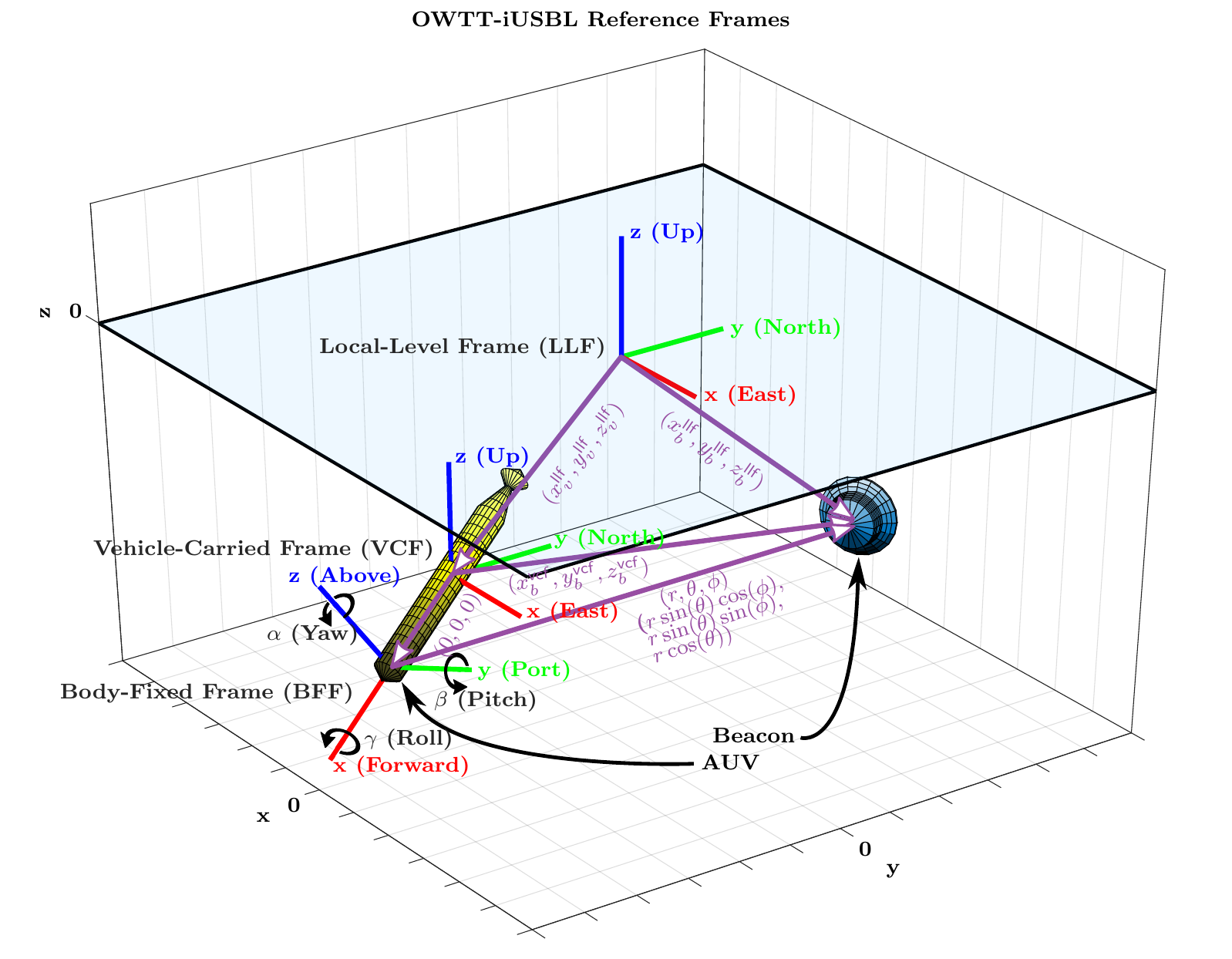}
    \caption{OWTT-iUSBL reference frames.}
    \label{fig:reference_frames}
\end{figure}

Acoustic processing of data measured by the OWTT-iUSBL receiver provides measurement distributions over range and azimuth and inclination.  Combining the MLE values from these acoustic measurements with an estimate of platform attitude from the IMU and magnetometer provides an instantaneous estimate of the relative $(x,y,z)$ position between the beacon and the vehicle;  however, such an approach is especially problematic in underwater acoustic navigation since sound propagation is prone to undesirable effects such as multi-path, interference and reflections, leading to false maxima -- this is evident in the range measurements of figure~\ref{fig:acoustic_measurement_distributions}, which illustrates how the use of MLE values will result in measurement outliers.  The multimodal nature of the acoustic range and angle measurements, as well as the fact that the projection of angular distributions into Cartesian coordinates in Euclidean space result in non-linear distributions, motivated the use of a particle filter for accurately capturing the distribution over the vehicle state estimate, rather than conventional parametric approaches such as the extended kalman filter (EKF).  This particle filter is used by each AUV to estimate the relative position of the beacon, by fusing attitude and heading measurements from the IMU and magnetometer, speed estimates from propeller RPM, and acoustic range and angle measurements from the OWTT-iUSBL receiver, and runs on-board the vehicle in real-time for online navigation.

\subsubsection{Reference frames, state description and transformations}

The various reference frames used by the particle filter are illustrated in figure~\ref{fig:reference_frames}.  The three relevant frames of reference are: the local-level frame (LLF), which is locally tangential to the Earth's surface with East-North-Up (ENU) $x$-$y$-$z$-aligned axes; the vehicle-carried frame (VCF), which also follows the ENU convention, but whose origin is centered around the AUVs center-of-gravity; and the body-fixed frame (BFF) which rotates with the orientation of the vehicle, and whose origin, like the VCF, is the vehicle's center-of-gravity (the BFF is shown offset from the VCF at the nose of the AUV in figure~\ref{fig:reference_frames} for clarity).  The $x$-$y$-$z$ axes of the body-fixed frame (BFF) are aligned with the vehicle forward, port and above directions, with roll ($\gamma$), pitch ($\beta$) and yaw ($\alpha$) following the right-hand rotation rule -- complete alignment of the $x$-$y$-$z$ axes of the BFF and VCF corresponds to zero rotation in all three axes, with the vehicle level and pointing toward the East.  Acoustic measurements in range ($r$), and angle (azimuth $\phi$ and inclination $\theta$) occur in the BFF, as shown in figure~\ref{fig:reference_frames}, with zero azimuth pointing along the x-axis and rotating $360^\circ$ with yaw, and zero inclination pointing along the z-axis and rotating $180^\circ$ with pitch.

The particle filter is used to track the position of the beacon relative to the AUV (i.e. the beacon position in the vehicle-carried frame), and partially tracks the state vector given by:

\begin{align}
\boldsymbol{\mathsf{x}} (t) &=
\begin{bmatrix}
\boldsymbol{x}_{b}^{\mathsf{vcf}} (t)^\intercal,
\boldsymbol{x}_{b}^{\mathsf{llf}} (t)^\intercal,
\boldsymbol{x}_{v}^{\mathsf{llf}} (t)^\intercal,
\boldsymbol{\gamma}_{v} (t)^\intercal,
\mathsf{sog}_{v}(t)
\end{bmatrix}^\intercal
\label{eq:state_vector}
\end{align}
\begin{align}
\boldsymbol{x}_{b}^{\mathsf{vcf}} (t) &=
\begin{bmatrix}
x_b^{\mathsf{vcf}} (t) \\[6pt]
y_b^{\mathsf{vcf}} (t) \\[6pt]
z_b^{\mathsf{vcf}} (t)
\end{bmatrix}
,\enspace
\boldsymbol{x}_{b}^{\mathsf{llf}} (t) =
\begin{bmatrix}
x_b^{\mathsf{llf}} (t) \\[6pt]
y_b^{\mathsf{llf}} (t) \\[6pt]
z_b^{\mathsf{llf}} (t)
\end{bmatrix}
,\enspace
\boldsymbol{x}_{v}^{\mathsf{llf}} (t) =
\begin{bmatrix}
x_v^{\mathsf{llf}} (t) \\[6pt]
y_v^{\mathsf{llf}} (t) \\[6pt]
z_v^{\mathsf{llf}} (t)
\end{bmatrix}
,\enspace
\boldsymbol{\gamma}_{v} (t) =
\begin{bmatrix}
\gamma_v (t) \\[6pt]
\beta_v (t) \\[6pt]
\alpha_v (t)
\end{bmatrix}
\label{eq:state_vector_detailed}
\end{align}
where $\boldsymbol{x}_{b}^{\mathsf{vcf}}$ and $\boldsymbol{x}_{b}^{\mathsf{llf}}$ are the beacon positions in the VCF and LLF respectively, $\boldsymbol{x}_{v}^{\mathsf{llf}}$ is the AUV position in the LLF, $\boldsymbol{\gamma}_{v}$ is the vector of AUV attitude Euler angles, and $\mathsf{sog}_{v}$ is the AUV speed-over-ground.  Only the beacon position in the VCF, $\boldsymbol{x}_{b}^{\mathsf{vcf}}$, is tracked by the particle filter, while other elements are either set constant prior to deployment or are values estimated from vehicle sensor measurements.

The transformation between the VCF and the BFF is simply given by the combination of the elemental rotations:

\begin{align}
&\boldsymbol{R}_{\mathsf{vcf}}^{\mathsf{bff}}(\boldsymbol{\gamma}_v) = \boldsymbol{R}_z(\alpha_v)\boldsymbol{R}_y(\beta_v)\boldsymbol{R}_x(\gamma_v)  \label{eq:rotation_transform}\\[6pt]
\boldsymbol{R}_z(\alpha_v) = 
\begin{bmatrix}
\cos\alpha_v & -\sin\alpha_v & 0 \\
\sin\alpha_v & \cos\alpha_v & 0 \\
0 & 0 & 1
\end{bmatrix},
\
&\boldsymbol{R}_y(\beta_v) = 
\begin{bmatrix}
\cos\beta_v & 0 & \sin\beta_v \\
0 & 1 & 0 \\
-\sin\beta_v & 0 & \cos\beta_v
\end{bmatrix},
\
\boldsymbol{R}_x(\gamma_v) = 
\begin{bmatrix}
1 & 0 & 0 \\
0 & \cos\gamma_v & -\sin\gamma_v \\
0 & \sin\gamma_v & \cos\gamma_v
\end{bmatrix}. \nonumber
\end{align}
Since acoustic angle and range measurements are determined using spherical coordinates within the BFF, transformations between spherical and Cartesian coordinates of the beacon in the BFF are given by the standard formulations:

\begin{align}
\boldsymbol{x}_{b}^{\mathsf{bff}} &=
\begin{bmatrix}
x_b^{\mathsf{bff}} \\[6pt]
y_b^{\mathsf{bff}} \\[6pt]
z_b^{\mathsf{bff}}
\end{bmatrix}
=
\begin{bmatrix}
r\sin\theta\cos\phi \\[6pt]
r\sin\theta\sin\phi \\[6pt]
r\cos\theta
\end{bmatrix},
\qquad
\begin{matrix}
r = || \boldsymbol{x}_{b}^{\mathsf{bff}} || = \sqrt{(x_b^{\mathsf{bff}})^2 + (y_b^{\mathsf{bff}})^2 + (z_b^{\mathsf{bff}})^2}, \\[6pt]
\theta = \arccos(z_b^{\mathsf{bff}}/r), \\[6pt]
\phi = \arctan(y_b^{\mathsf{bff}}/x_b^{\mathsf{bff}}).
\end{matrix} \label{eq:spherical_cartesian}
\end{align}
Finally, if the position of the beacon in the LLF is known ($\boldsymbol{x}_{b}^{\mathsf{llf}}$), vehicle position in the LLF ($\boldsymbol{x}_{v}^{\mathsf{llf}}$) can be determined via the particle filter's estimate of the beacon position in the VCF ($\boldsymbol{x}_{b}^{\mathsf{vcf}}$):

\begin{align}
\boldsymbol{x}_{v}^{\mathsf{llf}} = \boldsymbol{x}_{b}^{\mathsf{llf}} - \boldsymbol{x}_{b}^{\mathsf{vcf}}.
\label{eq:auv_llf}
\end{align}
In this work, our command, control and coordination strategy relies on a beacon-centric moving frame of reference; this beacon-centric frame is realized by fixing the beacon's LLF $x$-$y$ coordinates at the center of the local-level frame, with its $z$ coordinate at a known, constant beacon depth:

\begin{align}
\boldsymbol{x}_{b}^{\mathsf{llf}} =
\begin{bmatrix}
0, 0, -z_{b_0}^{\mathsf{llf}}
\end{bmatrix}^\intercal.
\end{align}
As a result, the $x$-$y$ position of the AUV in this moving, beacon-centric LLF is the negation of the particle filter estimate of the beacon position in the VCF, following equation~\ref{eq:auv_llf}; note that the $z$ coordinate of the vehicle in the LLF is measured directly using the AUV's pressure sensor.

\subsubsection{Initialization}

The particle filter maintains a set of $500$ particles, each of which holds a realization of the beacon position in the VCF with an associated weight:

\begin{align}
\mathsf{p}_x^{(i)} = \{\boldsymbol{x}_b^{\mathsf{vcf},(i)}, \enspace w_{\boldsymbol{x}}^{(i)}\} \qquad i=1,...,500.
\end{align}
The number of particles was chosen to ensure that the cycle frequency of the filter remained above the repeat rate of the beacon at $1$~\si{\hertz} -- one iteration of the filter takes about $500$~\si{\milli\second} on the Raspberry Pi 3 with $500$ particles.  These particles are initialized in a rather unorthodox manner -- two additional, separate sets of $500$ particles are first initialized, one in the range domain and the other in the angle domain using spherical coordinates:

\begin{align}
\mathsf{p}_r^{(i)} = \{r^{(i)}, \enspace w_r^{(i)}\}, \qquad \mathsf{p}_{\theta,\phi}^{(i)} = \{\theta^{(i)}, \enspace \phi^{(i)}, \enspace w_{\theta,\phi}^{(i)}\} \qquad i=1,...,500.
\end{align}
The range-domain particles $\mathsf{p}_r^{(i)}$ are initialized uniformly randomly between $0$~\si{\meter} and $316$~\si{\meter} (the maximum range of the OWTT-iUSBL system assuming a soundspeed of $1481$~\si{\meter\per\second}), while the angle-domain particles $\mathsf{p}_{\theta,\phi}^{(i)}$ are initialized according to a uniform distribution over the surface of a unit sphere; the particles in all sets are initialized with equal weighting of $\frac{1}{500}$.  A spherical to Cartesian coordinate transformation like that of equation~\ref{eq:spherical_cartesian} is then used to initialize the particles in the primary set, $\mathsf{p}_x^{(i)}$, resulting in the particles being initialized with a uniform random distribution within the volume of a sphere of radius $316$~\si{\meter}.  Besides initialization, particles in the secondary sets, $\mathsf{p}_r^{(i)}$ and $\mathsf{p}_{\theta,\phi}^{(i)}$, are used when incorporating acoustic measurements, as explained in section~\ref{sec:acoustic_update} later in this paper.  This initialization occurs whenever the AUV surfaces, since acoustic signal is lost and the beacon is free to move from its previous position.

\subsubsection{AUV motion model prediction}

The particle filter makes use of a simple constant velocity kinematics motion model for prediction:

\begin{align}
\boldsymbol{x}_{b}^{\mathsf{vcf},(i)} (t + \Delta t) &=
\begin{bmatrix}
x_b^{\mathsf{vcf},(i)} (t + \Delta t) \\[6pt]
y_b^{\mathsf{vcf},(i)} (t + \Delta t) \\[6pt]
z_b^{\mathsf{vcf},(i)} (t + \Delta t)
\end{bmatrix}
=
\begin{bmatrix}
x_b^{\mathsf{vcf},(i)} (t) - (\mathsf{sog}_{v}(t) + \mathcal{N}(0,\sigma_{\mathsf{sog}}^2)) \Delta t \sin(\alpha_v (t) + \mathcal{N}(0,\sigma_{\alpha}^2)) \\[6pt]
y_b^{\mathsf{vcf},(i)} (t) - (\mathsf{sog}_{v}(t) + \mathcal{N}(0,\sigma_{\mathsf{sog}}^2)) \Delta t \cos(\alpha_v (t) + \mathcal{N}(0,\sigma_{\alpha}^2)) \\[6pt]
z_b^{\mathsf{vcf},(i)} (t) - (z_v^{\mathsf{llf}} (t + \Delta t) - z_v^{\mathsf{llf}} (t))
\end{bmatrix}
\end{align}
where individual particles are propagated in the $x$-$y$ plane using AUV speed-over-ground estimated via equation~\ref{eq:speed_over_ground} and heading from the IMU, each modeled using Gaussian noise with specified standard deviations $\sigma_{\mathsf{sog}}$ and $\sigma_{\alpha}$; changes in $z$ are informed solely from changes in AUV depth as measured by its pressure sensor.  In addition, Gaussian process noise proportional to the expected maximum speed of the beacon is added to each particle's $x$-$y$ position to account for beacon movement.   Note that if our AUVs possessed a sensor such as a DVL to measure water velocity or speed-over-ground directly, the above motion model could be augmented to include these sensor measurements and improve the model prediction.

\subsubsection{Acoustic measurement update} \label{sec:acoustic_update}

Predictions are corrected by fusing acoustic range and angle measurements during the filter update step.  Ideally, the particles in the filter would cover the entire measurement space, represented by the volume of a  $316$~\si{\meter} radius sphere, with enough density to be able to capture the details of the measurement distribution; unfortunately, this is not feasible in practice, as it requires a very large number of particles.  To achieve computational tractability while retaining the high dynamic range of a filter with a much larger number of particles, we use a factored approach to incorporate range and angle measurements separately.  To begin the update step, the particles and their respective weights are first duplicated; one of these duplicate sets is transformed into the range domain:

\begin{align}
\mathsf{p}_r^{(i)} = \{r^{(i)}, \enspace w_r^{(i)}\} \quad \textsc{where} \quad \enspace r^{(i)} = || \boldsymbol{x}_{b}^{\mathsf{vcf},(i)} ||, \enspace w_r^{(i)} = w_{\boldsymbol{x}}^{(i)}
\end{align}
and the other duplicate set is transformed into the angle domain using equations~\ref{eq:rotation_transform} and \ref{eq:spherical_cartesian}:

\begin{align}
\begin{matrix}
\mathsf{p}_{\theta,\phi}^{(i)} = \{\theta^{(i)}, \enspace \phi^{(i)}, \enspace w_{\theta,\phi}^{(i)}\} \\[6pt]
\textsc{where} \\[6pt]
\boldsymbol{x}_{b}^{\mathsf{bff},(i)} = \boldsymbol{R}_{\mathsf{vcf}}^{\mathsf{bff}}(\boldsymbol{\gamma}_v) \cdot \boldsymbol{x}_{b}^{\mathsf{vcf},(i)}, \\[6pt]
\theta^{(i)} = \arccos(z_b^{\mathsf{bff},(i)}/r^{(i)}), \\[6pt]
\phi^{(i)} = \arctan(y_b^{\mathsf{bff},(i)}/x_b^{\mathsf{bff},(i)}), \\[6pt]
w_{\theta,\phi}^{(i)} = w_{\boldsymbol{x}}^{(i)}.
\end{matrix} \label{eq:angle_domain_transform}
\end{align}
To incorporate the acoustic range measurement, the weights of the particles in the range-domain set are multiplied with the matched filter output evaluated using equations~\ref{eq:II_phat}--\ref{eq:II_normalizematchedfilter}, at the nearest corresponding range.

A close coupling of sensor pair decomposition beamforming (SPD-BF) and the factored particle filter is used to incorporate the acoustic angle measurement -- rather than suffer the computational expense of evaluating the SPD beamformer over a fixed set of azimuths and inclinations, SPD-BF is performed only at the azimuths and inclinations represented by the particles in the angle-domain set, $\mathsf{p}_{\theta,\phi}^{(i)}$; these particles are then re-weighted by multiplying their corresponding weights with the result of these evaluations.  This close coupling enables the particle filter to run in real-time on the payload computer at a rate of approximately $2$~\si{\hertz}.  Note that since equation~\ref{eq:angle_domain_transform} involves a rotational transform using AUV attitude, it is important that this attitude reflects the state of the vehicle at the time at which the signal was received -- to ensure this, we maintain a rolling buffer of AUV attitude over the past second, and extract the pitch, roll and yaw values at the time closest to the maximum of the range measurement, and use these values as the input to this transformation.

The weights of both particle sets are then each re-normalized to sum to one, and the sets are reordered in ascending order according to their weights.  Finally, in order to obtain the updated primary particle set in the VCF, the range-domain and angle-domain sets are combined through element-wise multiplication and through transformations from spherical coordinates in the BFF to Cartesian coordinates in the VCF:

\begin{align}
\boldsymbol{x}_{b}^{\mathsf{vcf},(i)} &=
\begin{bmatrix}
x_b^{\mathsf{vcf},(i)} \\[6pt]
y_b^{\mathsf{vcf},(i)} \\[6pt]
z_b^{\mathsf{vcf},(i)}
\end{bmatrix}
=
\boldsymbol{R}_{\mathsf{vcf}}^{\mathsf{bff}}(\boldsymbol{\gamma}_v)^\intercal \cdot
\begin{bmatrix}
r^{(i)}\sin\theta^{(i)}\cos\phi^{(i)} \\[6pt]
r^{(i)}\sin\theta^{(i)}\sin\phi^{(i)} \\[6pt]
r^{(i)}\cos\theta^{(i)}
\end{bmatrix}, \qquad
w_{\boldsymbol{x}}^{(i)} = \frac{w_{r}^{(i)} w_{\theta,\phi}^{(i)}}{\sum_1^{500} w_{r}^{(i)} w_{\theta,\phi}^{(i)}} .
\end{align}
This factored particle filter approach allows us to approximate the behavior of a conventional particle filter operating in the full range-angle measurement domain using significantly fewer particles; since highly-weighted and lowly-weighted particles in the range and angle sets are associated through reordering, our approach effectively samples the dynamic range of a much larger set of particles in a conventional filter.  Although not strictly correct, this factored approximation works very well in practice.

At every update step particle weights are reevaluated and multiplied against the measurement distributions -- the particle weight $w_{\boldsymbol{x}}^{(i)}$ represents the likelihood of the particle's state hypothesis $\boldsymbol{x}_{b}^{\mathsf{vcf},(i)}$.  Consequently, after a string of updates, the distribution of weights across all particles can become uneven, with a few highly-weighted particles (strong hypotheses) and a majority with very low weight (weak hypotheses).  This is problematic, as it can drastically bias the filter's state estimate and reduce its performance.  We prevent degradation of filter performance by carrying out a resampling step after the update step and before the next predict step, in order to substitute single highly-weighted particles with a number of equally-weighted particles so as to maintain a more evenly-weighted particle set.  We adopt the systematic resampling scheme, due to its computational simplicity and empirically-demonstrated good performance \citep{Douc2005}.  In addition, to prevent the filter from converging prematurely to an incorrect mode, the $50$ lowest-weighted particles in both the range and angle domain sets are uniformly re-initialized over their respective domains before recombination, allowing the filter to re-converge onto the correct mode if premature convergence occurs.

\subsubsection{Likelihood estimation}

\begin{figure}[!t]
    \centering
    \includegraphics[width=1\textwidth]{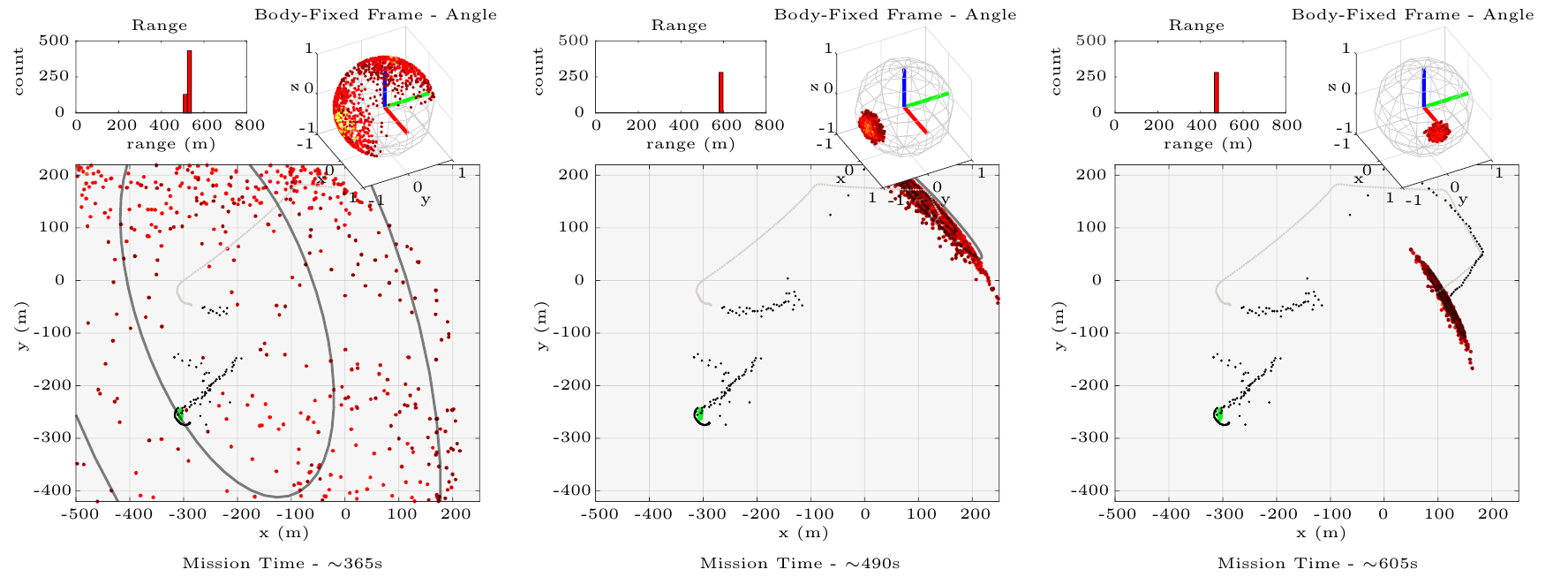}
    \caption{Evolution of the OWTT-iUSBL factored particle filter over time -- acoustic angle and range measurements are incorporated using duplicate sets of particles, $\mathsf{p}_{\theta,\phi}^{(i)}$ (top right subplot) and $\mathsf{p}_r^{(i)}$ (histogram in top left subplot), in the body-fixed frame (BFF), then re-combined and transformed into the vehicle-carried frame (VCF) to update particles in the primary set, $\mathsf{p}_x^{(i)}$.  The main axes show particles (colored red--yellow, with yellow having higher weight) in the local-level frame (LLF), calculated by offsetting VCF particles by beacon position (shown in green) in post-processing.  The particle filter estimate is shown in black, along with fitted $1\sigma$ and $2\sigma$ ellipses, as well as true AUV position using DVL shown in grey; note that the $50$ lowest-weighted particles used to prevent premature convergence are not visualized.}
    \label{fig:filter_evolution}
\end{figure}

The set of weighted particles $\mathsf{p}_x^{(i)}$ in the vehicle-carried frame (VCF) represent the probability distribution over the position of the acoustic beacon relative to the AUV, and must be converted into a single Cartesian coordinate for the purposes of vehicle control.  This is achieved through the calculation of the weighted mean over all particles, with the weighted covariance providing a measure of confidence in the estimate:

\begin{align}
\begin{matrix}
\tilde{\boldsymbol{x}}_{b}^{\mathsf{vcf}}(t) = 
\begin{bmatrix}
\tilde{x}_{b}^{\mathsf{vcf}}(t) \\[6pt]
\tilde{y}_{b}^{\mathsf{vcf}}(t) \\[6pt]
\tilde{z}_{b}^{\mathsf{vcf}}(t)
\end{bmatrix}
=
\begin{bmatrix}
\sum_{i=1}^{500} w_{\boldsymbol{x}}^{(i)} x_{b}^{\mathsf{vcf},(i)}(t) \\[6pt]
\sum_{i=1}^{500} w_{\boldsymbol{x}}^{(i)} y_{b}^{\mathsf{vcf},(i)}(t) \\[6pt]
z_b^{\mathsf{llf}} (t) - z_v^{\mathsf{llf}} (t)
\end{bmatrix} 
\vspace{12pt}\\
\boldsymbol{\Sigma}_{b}^{\mathsf{vcf}}(t) = \frac{1}{499} \sum_{i=1}^{500} w_{\boldsymbol{x}}^{(i)}
\begin{bmatrix}
x_{b}^{\mathsf{vcf},(i)}(t) - \tilde{x}_{b}^{\mathsf{vcf}}(t) \\[6pt]
y_{b}^{\mathsf{vcf},(i)}(t) - \tilde{y}_{b}^{\mathsf{vcf}}(t)
\end{bmatrix}
\begin{bmatrix}
x_{b}^{\mathsf{vcf},(i)}(t) - \tilde{x}_{b}^{\mathsf{vcf}}(t), \enspace
y_{b}^{\mathsf{vcf},(i)}(t) - \tilde{y}_{b}^{\mathsf{vcf}}(t)
\end{bmatrix}
\end{matrix}.
\end{align}
Note that the particles in the filter are used only to estimate the position of the acoustic beacon relative to the AUV in the $x$ and $y$ coordinates of the VCF, while the known depth of the beacon and the depth of the vehicle from pressure sensor measurements are used to estimate the beacon's $z$ coordinate.

Figure~\ref{fig:filter_evolution} illustrates how the factored particle filter evolves over time, during a mission in which an AUV detects, localizes, then drives toward the acoustic beacon using the OWTT-iUSBL navigation system.  In this case, the OWTT-iUSBL system was configured for a larger range of $800$~\si{\meter}, and fielded on a Bluefin-21 AUV outfitted with a tetrahedral nose array and a DVL for ground-truth validation.  In the first image of figure~\ref{fig:filter_evolution}, it can be seen that the particles (shown as red--yellow colored dots) have clustered into two groups in the angle domain, representing the main lobe and sidelobe of the acoustic beamforming output; as a result, the filter has two modes, and has yet to converge.  In the second image, the filter has converged onto the correct mode, and by the third image it is tracking this mode and providing a good estimate of vehicle position.

\section{Command, control and coordination of multiple AUVs with relative acoustic navigation}\label{sec:relative_acoustic_navigation}

\begin{figure}[!t]
    \centering
    \includegraphics[width=1\textwidth]{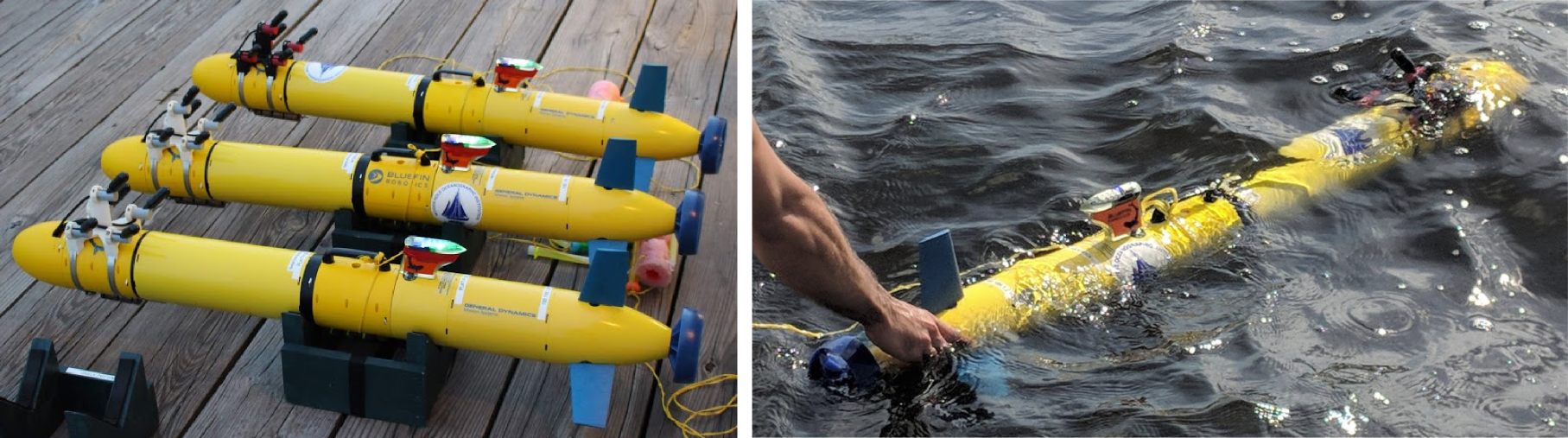}
    \caption{Photograph of the fleet of three Bluefin SandShark AUVs, named \emph{Platypus}, \emph{Wombat} and \emph{Quokka} (left), outfitted with the OWTT-iUSBL payload; \emph{Quokka} being hand-deployed dockside (right).}
    \label{fig:fleet_photos}
\end{figure}

The one-way travel-time inverted ultra-short baseline (OWTT-iUSBL) system detailed in the prior section provides an acoustic navigation solution for autonomous underwater vehicles (AUVs) that carries with it two important qualities that make it especially suited for multi-vehicle applications: firstly, OWTT ranging ensures system scalability, since any number of AUVs within range are able to localize themselves by passively receiving the beacon signal; and secondly, the system is suitably inexpensive and low-power (since the system does not transmit acoustically) to be used on the emerging class of miniature, low-cost AUV that has neither the space nor the energy density to make use of standard AUV navigational sensors such as the DVL.  In this section, we elaborate on a third benefit of the OWTT-iUSBL system for multi-AUV deployments -- the use of a single navigation beacon, and the operation of vehicles which navigate themselves relative to that beacon, enables an intuitive operating paradigm which allows a single operator to direct and control the behavior of the entire vehicle fleet.  In this work, we demonstrate the capability of this paradigm to command, control and coordinate a fleet of three SandShark AUVs, named \emph{Platypus}, \emph{Quokka} and \emph{Wombat}, pictured in figure~\ref{fig:fleet_photos}.

\subsection{Coordination with relative autonomy} \label{sec:behaviors}

\begin{figure}[!t]
    \centering
    \includegraphics[width=1\textwidth]{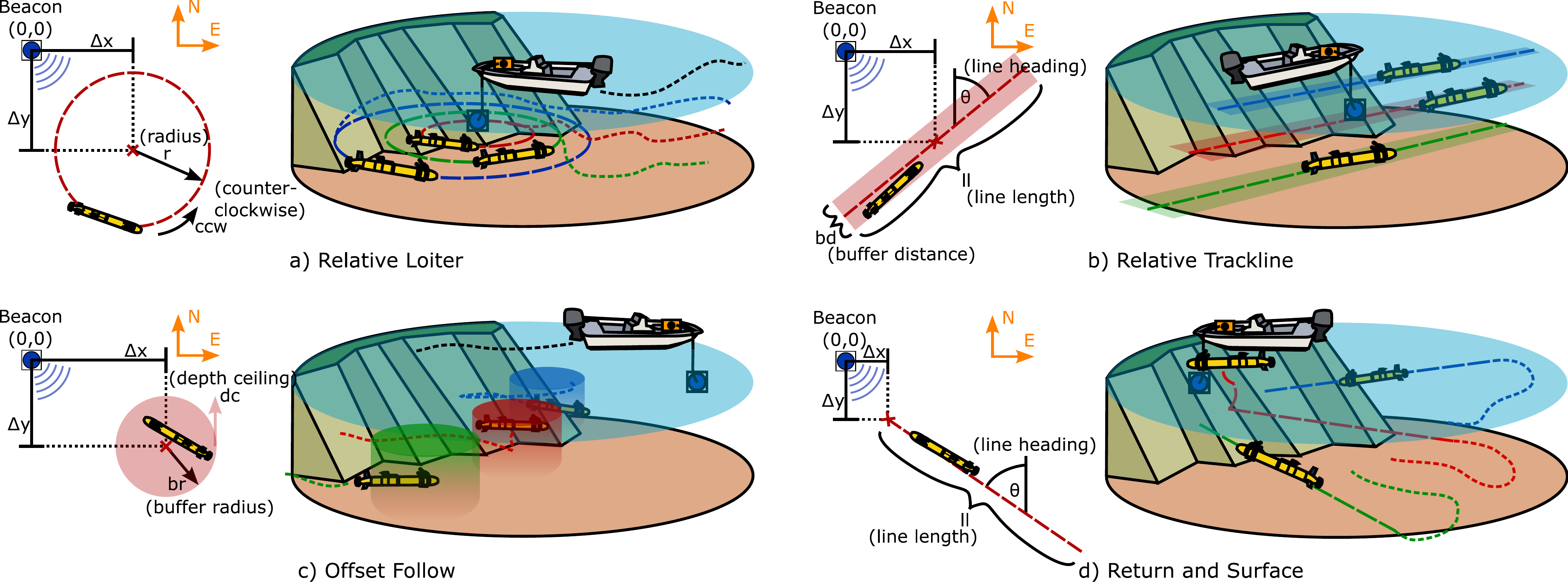}
    \caption{Illustration of autonomous AUV behaviors designed for beacon-relative acoustic navigation; a) the relative loiter has vehicles track circular trajectories at different radii and/or offsets from the beacon; b) the relative trackline has each vehicle follow a trackline at different beacon offsets and headings and can be used to survey an area cooperatively; c) the offset follow behavior has vehicles attempt to maintain a waypoint offset to the beacon within a depth cylinder for cooperative formation keeping; d) return and surface has the vehicles home in on the beacon at different headings and surface at the end of a mission.}
    \label{fig:relative_behaviors}
\end{figure}

Planned trajectories for autonomous AUV behaviors such as waypoints, racetracks, loiters and lawnmower paths are typically defined within an absolute frame of reference.  When the OWTT-iUSBL beacon is static at a known position, or if it is able to transmit its absolute position within its broadcast signal, then vehicles would be able to self-localize in the absolute reference frame and these conventional behaviors can be used to coordinate vehicles.  However, in our case we are interested in coordinating multiple AUVs without them knowing the absolute beacon position -- instead, we design vehicle behaviors such that they operate relative to the acoustic beacon, in a beacon-centric moving frame of reference.  These behaviors are implemented using the MOOS-IvP autonomy framework \citep{Benjamin2010}; and behavior parameters are set for each individual vehicle in the context of the fleet, such that coordination between multiple AUVs is achieved.  In this work, we present initial investigations into this approach using four autonomous behaviors designed to operate within this beacon-relative navigation paradigm; these behaviors are illustrated in figure~\ref{fig:relative_behaviors}.

The first of these behaviors is the relative loiter.  This behavior simply directs the vehicle to track a circular trajectory of a specified radius in a clockwise or counterclockwise direction, centered at specified $x$ and $y$ standoff distances from the beacon; if the standoffs are both set to zero, this corresponds to the vehicle continuously circling the beacon.

The second behavior, the relative trackline, is designed to command the vehicle to travel along a finite-length line transect along a specified heading.  The center of this line is positioned at a user-defined $x$ and $y$ offset from the beacon, with the length and heading of the line determined by additional parameters.  If the beacon is static, the resulting behavior of the AUV is to continuously travel back-and-forth, tracking this line; however, if the beacon is moved in a direction perpendicular to the trackline, the operator is able to sweep the line over an area, consequently causing the vehicle to automatically survey that area with successive transects.  Another parameter controls the width of a buffer region around the trackline -- this region prevents oscillatory AUV behavior due to imperfect estimation of the relative beacon position, by directing the vehicle to center itself along the trackline only when it has drifted outside this buffer.

The offset follow is a simple waypoint-like behavior, where the vehicle always attempts to maintain an $x$ and $y$ offset from the beacon.  Upon reaching this relative position, the AUV stops its propeller, and given its slightly positive buoyancy, slowly floats toward the surface; if the vehicle floats above a specified depth ceiling, the AUV thrusts again and circles back to the desired position at depth.  As with the relative trackline, a radius parameter defines a buffer region within which the vehicle propeller will remain inactive if the AUV has reached position -- if the vehicle drifts, or if the beacon moves such that the AUV exits this region, then the vehicle once again drives to maintain the offset at depth.  The result is a sprint-and-drift behavior where AUVs can be directed to maintain a coordinated formation using vehicle-specific $x$ and $y$ beacon offsets.

The final behavior is the return and surface behavior, which is formulated as a convenient method of recalling all vehicles in the fleet back to the operator at the beacon.  A finite-length return trackline along a user-specified heading ending at a given offset from the beacon is defined, the vehicle is directed along this line starting at the nearest point of approach, and upon reaching the end, stops all activity and floats to the surface for retrieval.

For each of these behaviors, coordination between vehicles in a fleet is achieved by setting behavior parameters that are specific to each AUV prior to launch -- individual vehicles set with different radii and offsets for the relative loiter, different offsets for the relative trackline and offset follow behaviors, and different headings for the return and surface behavior, provide a method to perform coordinated surveys, formation keeping and fleet retrieval while minimizing the risk of collisions, without the need for inter-vehicle communication.

\subsection{Command and behavior switching}

\begin{figure}[!t] 
	\centering
	\includegraphics[width=1\textwidth]{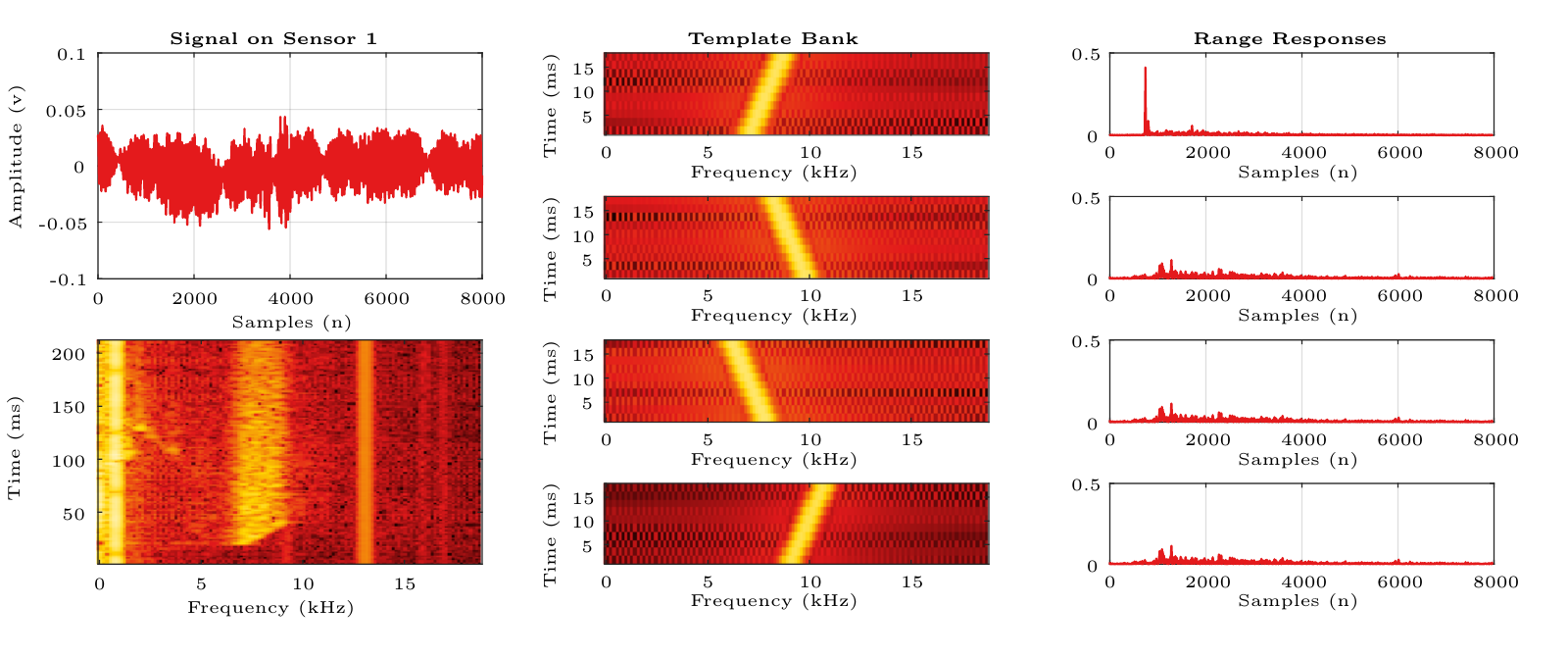}
	\caption{Visualization of mode identification on-board the AUV -- in-water signals recorded by the OWTT-iUSBL receiver array (left, for sensor $1$) are range processed using matched filtering against a bank of possible signal replicas (center, replicas for modes $1$, $2$, $3$ and $4$, top to bottom); range responses (right) are examined to find the largest response, which the vehicle identifies as the broadcast signal (mode $1$ in this case).} 
\label{fig:filter_bank}
\end{figure}

Operator command of the fleet is realized using a simple acoustic communications scheme.  Prior to deployment, autonomous behaviors on each vehicle are assigned to four different modes.  The operator is able to switch between these four modes by manually selecting one of four acoustic signals to be broadcast by the navigation beacon, using the beacon's 5-position rotary switch.  Once in the water, the AUVs detect and identify the broadcast signal, and perform the behavior associated with its corresponding mode, thereby carrying out the desired behavior as commanded by the operator.  The four acoustic signals corresponding to the four modes are illustrated in the middle of figure~\ref{fig:filter_bank}, and comprise of four linear frequency-modulated (LFM) $20$~\si{\milli\second} chirps at $7$-$9$~\si{\kilo\hertz}, $10$-$8$~\si{\kilo\hertz}, $8$-$6$~\si{\kilo\hertz} and $9$-$11$~\si{\kilo\hertz}.  Although this one-way communications scheme means that the operator receives no acknowledgment of correct mode determination by each vehicle, since the beacon continuously broadcasts the commanded signal for OWTT-iUSBL navigation, vehicles are almost guaranteed to undertake the correct behavior as long as the mode is commanded for a significant period of time ($10$s of seconds or more) and the vehicle is within receive range.  Indeed, during our experiments, all three vehicles always identified the correct mode within a few seconds of each other.

Figure~\ref{fig:filter_bank} shows an example of this mode identification process -- each AUV holds a template bank containing replicas of the four possible broadcast signals and their associated modes; the matched filtering range response of the AUV's array-recorded signals is evaluated using equations~\ref{eq:II_phat}--\ref{eq:II_normalizematchedfilter} against each of these replicas, and the replica that elicits the largest response is selected as the most likely broadcast signal; if the same replica is selected three times in succession, then the AUV determines that the mode associated with that replica has been commanded by the operator, and the vehicle performs the corresponding autonomous behavior.  Beamforming is also performed within the frequency band of the selected replica to facilitate OWTT-iUSBL localization -- note that beamforming over the wide bandwidth represented by all signals in the template bank is only made possible by the low memory requirements of the SPD-BF algorithm.  In the case of figure~\ref{fig:filter_bank}, the $7$-$9$~\si{\kilo\hertz} LFM chirp has clearly produced the largest response, corresponding to mode $1$.

\subsection{Implicit control with beacon movement} \label{sec:control}

\begin{figure}[!t]
    \centering
    \includegraphics[width=0.8\textwidth]{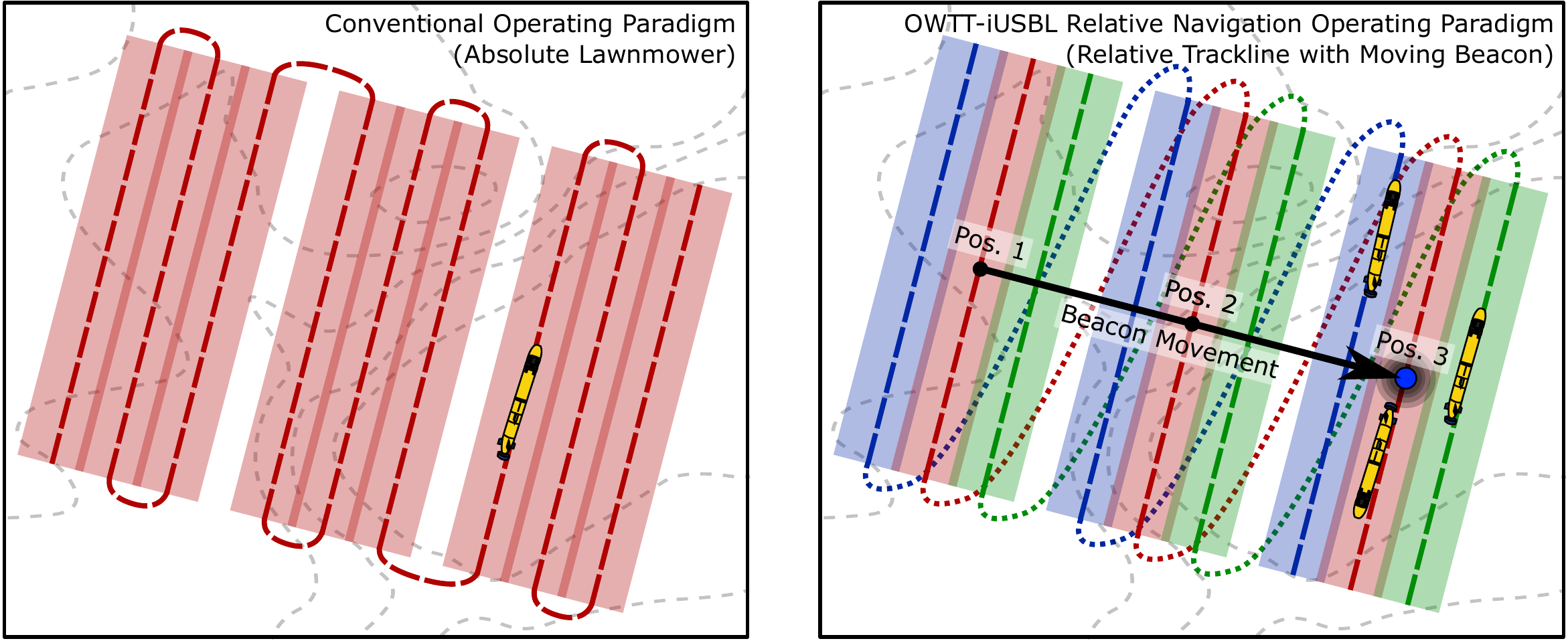}
    \caption{Conceptual illustration of the difference between a survey using the conventional and the relative acoustic navigation AUV operating paradigms -- conventional AUV operations require the survey to be programmed in the absolute reference frame (left), while relative acoustic navigation has each vehicle maintain the desired trajectory relative to the navigation beacon, allowing the same area to be surveyed through absolute re-positioning of the beacon (right).}
    \label{fig:fleet_control}
\end{figure}

Operator control of the AUV fleet is an inherent property of the relative acoustic navigation paradigm -- because AUVs operate within the beacon-centric frame of reference, absolute movement of the beacon itself results in a re-positioning of the entire fleet, providing the operator with a centralized means of controlling absolute fleet position.  This property is conceptually illustrated in figure~\ref{fig:fleet_control} --  in the conventional operating paradigm, surveying an area using a single AUV consists of planning and performing a vehicle path such as a lawnmower pattern in an absolute reference frame; alternatively, using multiple AUVs and the relative trackline behavior, the same area can be surveyed by the movement of the beacon itself, as shown on the right of figure~\ref{fig:fleet_control}.  As the operator re-positions the beacon at three different locations, successive tracklines by the vehicles as they maintain their programmed line offsets from the beacon results in the desired area coverage.  By recording the absolute position of the beacon, vehicle trajectories can be transformed into the absolute frame of reference for geolocalization of measurements in post-processing.

Although conceptually simple, this approach is powerful when combined with command broadcasts as detailed in the previous section.  In conventional multi-AUV operations, selection of behaviors, vehicle re-positioning, or even abort commands must be commanded by the transmission of individual, vehicle-specific acoustic requests, with the associated high latency that may be unacceptable in time-critical situations and in an approach that becomes more cumbersome with larger vehicle fleets.  In contrast, multi-AUV operations are considerably simplified with our relative navigation paradigm, by allowing a single operator to intuitively command and control fleet-wide behavior and position, regardless of the number of vehicles -- re-positioning vehicles to avoid hazardous areas or unexpected obstacles, aborting the mission, or requesting all AUVs to return to the operator are all easily accomplished using our approach.

\section{Multi-AUV river experiments}\label{sec:experiments}

\begin{figure}[!t]
    \centering
    \includegraphics[width=1\textwidth]{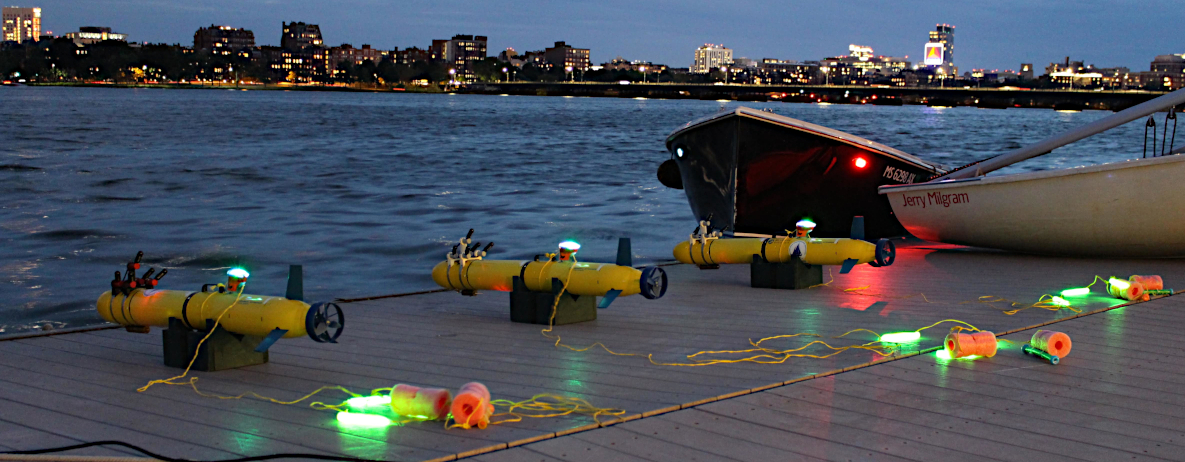}
    \caption{Photo of our Bluefin Robotics SandShark fleet during night operations in the Charles River with deployment from the MIT Sailing Pavilion and the Boston skyline in the background -- strobe lights in the AUV masts and towed buoys with glow sticks provided a convenient way to monitor the AUVs when underway and underwater; the motorboat tied to the dock was used by two operators to re-position the OWTT-iUSBL acoustic beacon and to broadcast different commands to the fleet.}
    \label{fig:fleet_photo_night}
\end{figure}

The one-way travel-time inverted ultra-short baseline (OWTT-iUSBL) positioning system, in conjunction with the relative acoustic navigation operating paradigm, was field tested using our fleet of three Bluefin Robotics SandShark AUVs over a total of six deployments in the Charles River in Boston during a period of three days in September 2018.  These deployments and prior tests of the system were performed in a segment of the river adjacent to the Massachusetts Institute of Technology (MIT) Sailing Pavilion, as pictured in figure~\ref{fig:fleet_photo_night}.  In this section we present results from these deployments, which represent the final performance of the system achieved after a number of weeks of system calibration, troubleshooting and refinement of vehicle behaviors, and tens of hours of multi-AUV operational experience.

\subsection{Preliminary experiments}\label{sec:prelim_experiments}

Preliminary experimental work and refinement of the OWTT-iUSBL relative navigation system in the Summer of 2018 was crucial for achieving the successful multi-AUV deployments presented in this paper.  Much of this work involved system testing using a single SandShark AUV, \emph{Platypus}, to debug our designed autonomous behaviors from section~\ref{sec:behaviors}, to refine control issues, and to investigate and calibrate sensor biases.  Here we elaborate on some lessons learned during this process.

An important insight we gained from repeated deployments of these small, lightweight vehicles is that they are notably more sensitive to changes in ballast, payload mass and profile, as compared to more conventional mid-sized or large AUVs.  This may not be surprising, given that even small changes to the payload, such as the addition or re-positioning of sensors, contribute a larger difference to the structural and mass properties of a smaller vehicle; as a result, proper ballasting and tuning of the control system is critical not only for effective vehicle motion, but also for the accuracy of OWTT-iUSBL localization.  In our case, we utilized AUV system identification trajectories, including separate runs with step-changes in depth, heading, and pitch, to tune vehicle PID gains over a period of three days.  OWTT-iUSBL localization accuracy is largely improved with smooth attitude and heading changes resulting from good vehicle control -- this is because the projection of angular measurement into Cartesian position is reliant on vehicle orientation estimates, which degrade when the AUV experiences sudden changes in attitude; this can result in position biases that are difficult to compensate for.  Fortunately, since all three vehicles are nearly identical, once one vehicle was properly tuned, the application of its PID gains to the other two vehicles achieved similar control performance.

Perhaps the most critical lesson learned during this process is the importance of acoustic calibration of the OWTT-iUSBL receiver.  This calibration must be done with the receiver installed on the AUV, since local acoustic interactions with the body of the vehicle produce biases in the acoustic angle measurement that are dependent on the azimuth and inclination to the acoustic source, as well as signal frequency.  Using the calibration process described at the end of section~\ref{sec:angle_measurement}, we observed that the azimuth-dependent bias of the angle measurement was stable and repeatable, with patterns identical across all three AUVs -- consequently, we compensated for these biases with a look-up table of this dependency, which significantly improved OWTT-iUSBL localization accuracy.  Additionally, the acoustic calibration process performed at different ranges allowed us to fit an effective soundspeed to collected OWTT measurements, with the soundspeed calculated to be approximately $1481$~\si{\meter\per\second}.

On-board compass calibration and evaluation also plays an important role in obtaining accurate localization estimates; one advantage of these small platforms is that calibration routines, which involve the collection of magnetometer measurements over the sphere, can be conveniently obtained by hand-rotating the AUV through various orientations.  Evaluation of heading accuracy of all three AUVs against a dual-antenna GPS determined that although not zero-mean, statistics were nicely Gaussian-distributed with a consistent $1\sigma$ of around $3^\circ$, and this observation allowed us to compensate for heading bias in the navigation filter.

Finally, preliminary deployments allowed us to test and refine the behaviors we designed for relative acoustic navigation; changes to behavior logic as a result of these experiments were essential in improving overall robustness -- one example is the inclusion of the buffer distance in the relative trackline behavior, which, as previously mentioned, helped remove oscillatory behavior that was observed in earlier iterations of the behavior.  The MIT Sailing Pavilion dock also enabled easy testing of the relative navigation paradigm, since the acoustic beacon could be hand-deployed into the water, and the operator could move the beacon up and down the length dock and observe the AUVs correctly following the beacon, as well as recall the vehicles back to the dock for easy retrieval.

\subsection{Multi-AUV experiments: 10\textsuperscript{th}, 12\textsuperscript{th} and 14\textsuperscript{th} September 2018}

The six deployments from which we gather data for the results presented here were each between $45$ and $65$ minutes long, and took place on the three days of the 10\textsuperscript{th}, 12\textsuperscript{th} and 14\textsuperscript{th} of September 2018.  These missions were performed over an area of about $300\times 200$~\si{\meter} in the section of the Charles River by the MIT Sailing Pavilion -- it should be noted that the river environment in this area is acoustically challenging, with shallow depths (minimum, average and maximum depths of about $3.9$~\si{\meter}, $5.3$~\si{\meter} and $6.5$~\si{\meter} respectively), a muddy bottom that is sound absorptive (limiting acoustic range), and an acoustically-reflective stone wall bounding the North side of the river (causing acoustic reflective interference in areas close to the dock).  Each of our three SandShark AUVs, \emph{Platypus}, \emph{Quokka} and \emph{Wombat} pictured in figure~\ref{fig:fleet_photo_night}, were programmed to dive and maintain a desired depth of $2.5$~\si{\meter}, and a speed-over-ground of $1$~\si{\meter\per\second} throughout these deployments; a pair of small polyethylene foam floats, tied to each vehicle with a few meters of floating line, were used by operators to monitor AUV behavior and their general position, since the vehicles had no capability for acoustic transmission.

The OWTT-iUSBL acoustic beacon was towed by a manually-piloted motorboat at a depth of approximately $1$~\si{\meter}; a passenger on the boat operated the rotary dial for mode selection and fleet command, while the pilot would drive and steer the motorboat to re-position the beacon within the absolute reference frame.  The boat carried with it a Hemisphere V102 dual-antenna GPS which allowed timestamped beacon position to be recorded with decimeter-level accuracy.

To validate OWTT-iUSBL localization data, two additional beacons were fastened to the MIT Sailing Pavilion dock  at about $1$~\si{\meter} depth; these beacons were located at approximately $(x,y) = (17.05, 1.78)$~\si{\meter} and $(x,y) = (-60.56, -34.97)$~\si{\meter} in the absolute frame of reference, providing an $85.87$~\si{\meter} baseline for a long-baseline (LBL) localization reference.  These beacons were \emph{not} used for AUV navigation -- however, they also broadcast an acoustic signal at the start of every second, in sync with OWTT-iUSBL beacon transmissions and recording on the AUV receivers.  LBL signals were outside the frequency band of the OWTT-iUSBL beacon, using $20$~\si{\milli\second} chirps at $5$-$2$~\si{\kilo\hertz} and $0.25$-$1.5$~\si{\kilo\hertz} for the East and West LBL beacons respectively.  Post-processing of acoustic recordings by each vehicle then allowed us to obtain AUV trajectories in the absolute reference frame using OWTT range intersections of the two LBL beacons -- because the operating area was limited to the South half of the LBL baseline, intersection ambiguity was easily broken.  Prior work by the authors \citep{Rypkema2019} demonstrated that this LBL setup in the same environment is able to provide positioning uncertainty with a standard deviation of $3.9$~\si{\meter}, comparable to a standard GPS receiver -- as such, we use this LBL system as ground-truth to validate OWTT-iUSBL navigation data.

\begin{figure}[!t]
    \centering
    \includegraphics[width=1\textwidth]{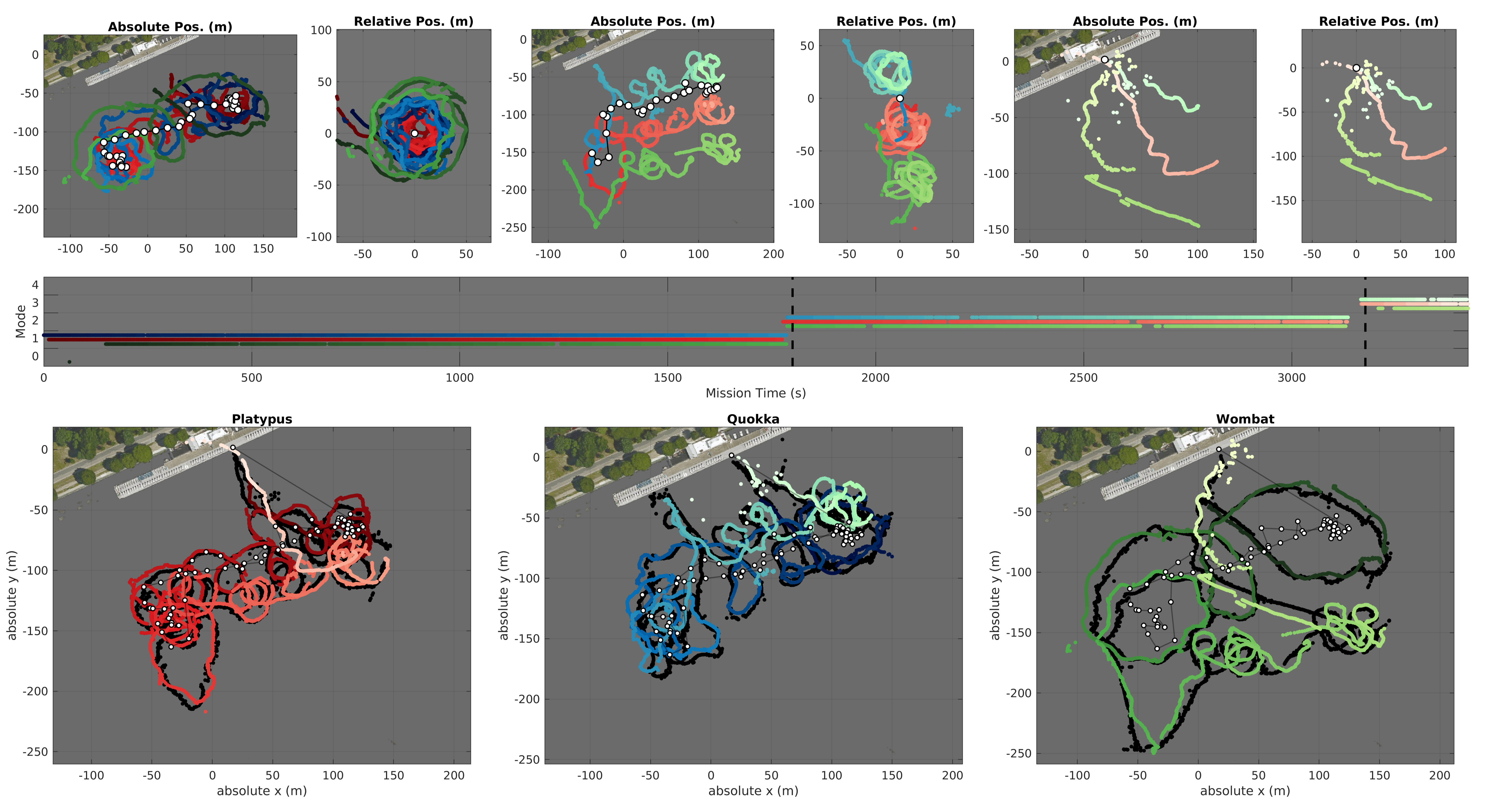}
    \caption{Mission 1 on September 10\textsuperscript{th} 2018, illustrating the relative loiter and return and surface behaviors for \emph{Platypus} (red), \emph{Quokka} (blue) and \emph{Wombat} (green), with darker-to-lighter shading indicating mission progression over time; the middle row shows the mode identified by each AUV over the mission duration; the top row includes AUV trajectories from OWTT-iUSBL navigation in the absolute and relative (beacon-centric) reference frames for the three mission periods as divided by the dashed black lines in the middle plot; the bottom row illustrates the full trajectory for each AUV separately, with OWTT-iUSBL navigation in color, LBL ground-truth position in black, and beacon position as connected white dots.}
    \label{fig:multi_experiment1}
\end{figure}

\subsubsection{Mission 1}
In this paper, we provide figures from the first and last of the six deployments to illustrate each of our four vehicle behaviors.  This first mission, which took place on September 10\textsuperscript{th} 2018, was configured such that modes $1$ and $2$ corresponded to relative loiters, mode $3$ was the return and surface behavior, and mode $4$ signaled an abort where all AUVs were commanded to stop all activity and float to the surface.  Mission parameters for each of these modes are listed in table~\ref{tbl:mission1} for each vehicle.

\begin{figure}[!t] 
	\centering
	\includegraphics[width=1\textwidth]{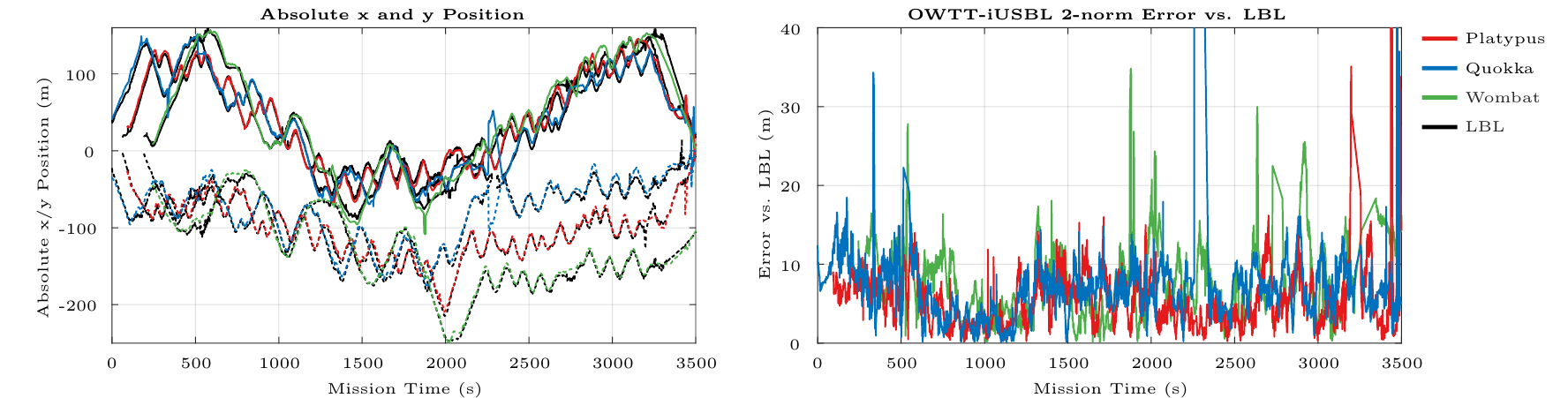}
	\caption{Mission 1 plots of $x$ (solid) and $y$ (dashed) positions from OWTT-iUSBL navigation in color, and LBL localization in black (left); $2$-norm error of OWTT-iUSBL against LBL ground-truth (right).} 
\label{fig:mission1_error}
\end{figure}

\begin{table}[b]
\caption{Modes and associated autonomous AUV behaviors and parameters for Mission 1.} \label{tbl:mission1}
\begin{center}
\begin{tabular}{|m{3cm}|c|c|c|c|}
  \hline
% after \\: \hline or \cline{col1-col2} \cline{col3-col4} ...
   \multicolumn{2}{|c|}{} & Platypus & Quokka & Wombat \\
  \hline\hline
  \multirow{3}{3cm}{Mode 1: \\Relative Loiter} & $(\Delta x, \Delta y)$ (m) & $(0,0)$ & $(0,0)$ & $(0,0)$ \\ \cline{2-5}
   & Radius (m) & $18$ & $36$ & $48$ \\ \cline{2-5}
   & Direction & CCW & CCW & CCW \\ \hline
  \multirow{3}{3cm}{Mode 2: \\Relative Loiter} & $(\Delta x, \Delta y)$ (m) & $(7.5,-26.0)$ & $(-7.5,26.0)$ & $(22.5,-78.0)$ \\ \cline{2-5}
   & Radius (m) & $18$ & $18$ & $18$ \\ \cline{2-5}
   & Direction & CCW & CCW & CCW \\ \hline
  \multirow{3}{3cm}{Mode 3: \\Return \& Surface} & $(\Delta x, \Delta y)$ (m) & $(0.0,-5.0)$ & $(2.2,-2.2)$ & $(-2.2,2.2)$ \\ \cline{2-5}
   & Line length (m) & $150$ & $150$ & $150$ \\ \cline{2-5}
   & Return heading ($^\circ$) & $340$ & $300$ & $20$ \\
  \hline
\end{tabular}
\end{center}
\end{table}

Trajectories of the fleet during this mission are plotted in figure~\ref{fig:multi_experiment1}.  The deployment began with the motorboat pilot and beacon operator driving out into the river and maintaining position around $(110,-70)$~\si{\meter}, and commanding mode $1$ from the beacon; a third operator at the dock then individually deployed \emph{Quokka}, \emph{Platypus} and \emph{Wombat}, with each vehicle successfully identifying the commanded mode (as shown in the middle plot of figure~\ref{fig:multi_experiment1}).  In this mode, the fleet undertook a counterclockwise relative loiter behavior centered at the beacon, with each AUV at different radii -- this mode lasted until around $1770$~\si{\second} into the mission, as the boat and the beacon drifted from East to West, ending around the position $(-40,-150)$~\si{\meter}.  During this period, the fleet correctly loitered around the beacon, as illustrated in the absolute and relative position plots at the top-left of figure~\ref{fig:multi_experiment1}; the relative position plot in particular clearly shows the AUVs loitering at their different programmed radii.  At around $1770$~\si{\second} the beacon operator switched to mode $2$, with this command identified correctly by the fleet.  In this mode, the fleet was programmed to maintain relative $18$~\si{\meter} counterclockwise loiters at different offsets from the beacon, spaced about $54$~\si{\meter} apart -- this mode lasted until around $3150$~\si{\second}, with the fleet successfully performing the desired behavior as the boat returned to its original position in the East, as shown in the top-middle plots of figure~\ref{fig:multi_experiment1}.  Finally, at around $3150$~\si{\second}, the operator on the boat switched off the beacon, and the East LBL beacon was set to transmit the signal to command mode $3$ by the dock operator, which resulted in all three AUVs returning to the dock at their commanded headings and surfacing for retrieval (top-right plots of figure~\ref{fig:multi_experiment1}).  Note that although we previously stated that the LBL beacons are \emph{not} used for navigation, the use of the East LBL beacon for this return and surface behavior is the singular exception, and done for the convenience of dockside vehicle retrieval.

The results of this mission plotted in figure~\ref{fig:multi_experiment1} clearly demonstrate the ability of our system to command, control and coordinate the AUV fleet, with each vehicle successfully responding to commanded modes and performing the desired behaviors, as well as maintaining their specified relative positions to the beacon as it moved in the absolute frame.  The bottom plots of figure~\ref{fig:multi_experiment1} show the OWTT-iUSBL trajectories of each AUV separately, transformed into the absolute reference frame by offsetting the relative trajectory calculated online by each AUV with the GPS position of the beacon (equation~\ref{eq:auv_llf}) in post-processing; comparison with LBL ground-truth in black shows good qualitative agreement between the two solutions.

The left of figure~\ref{fig:mission1_error} provides plots of vehicle $(x,y)$ position for all three AUVs, as calculated using our OWTT-iUSBL system, as well as from the LBL setup.  As with the trajectories in figure~\ref{fig:multi_experiment1}, it can be seen that OWTT-iUSBL navigation tracks the LBL ground-truth solution well;  the good agreement between the solutions is quantified by the the $2$-norm error of OWTT-iUSBL positioning referenced against LBL as plotted on the right of figure~\ref{fig:mission1_error}.  The position error for all vehicles generally fluctuates around $10$~\si{\meter} or less, with spikes that occur usually when the AUV gets near to the surface and obtains a GPS fix -- when this occurs the particle filter re-initializes, causing the particles to spread and the mean to be close to zero with a resulting incorrect solution close to the beacon.  Note also that the LBL system is not foolproof, and its solution is also subject to acoustic effects and small outliers; however, the similarity of trajectories from both systems provides strong support for the capacity of our OWTT-iUSBL system to provide accurate localization.  Although we do not illustrate the trajectories from dead-reckoning here, each vehicle also maintains a dead-reckoning solution internally which does not incorporate acoustic measurements -- error in dead-reckoning grows unbounded, and grew to a maximum of $50$--$120$~\si{\meter} during this mission for each AUV.

\begin{figure}[!t]
    \centering
    \includegraphics[width=1\textwidth]{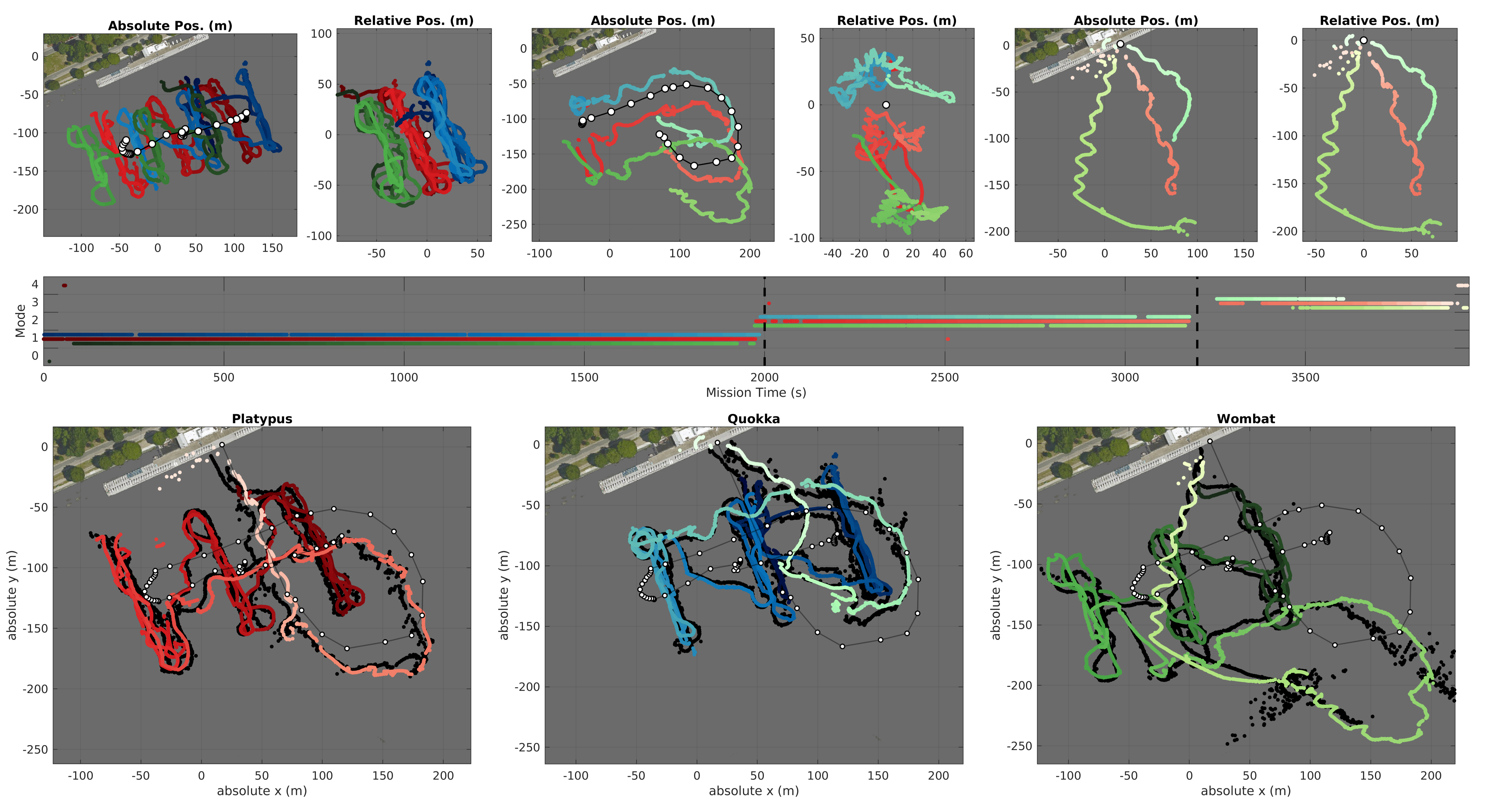}
    \caption{Mission 6 on September 14\textsuperscript{th} 2018, illustrating the relative trackline, offset follow, and return and surface behaviors for \emph{Platypus} (red), \emph{Quokka} (blue) and \emph{Wombat} (green), with darker-to-lighter shading indicating mission progression over time; the middle row shows the mode identified by each AUV over the mission duration; the top row includes AUV trajectories from OWTT-iUSBL navigation in the absolute and relative (beacon-centric) reference frames for the three mission periods as divided by the dashed black lines in the middle plot; the bottom row illustrates the full trajectory for each AUV, with OWTT-iUSBL navigation in color, LBL ground-truth position in black, and beacon position as connected white dots.}
    \label{fig:multi_experiment6}
\end{figure}

\subsubsection{Mission 6}
This sixth and final mission took place on September 14\textsuperscript{th} 2018, and was configured to execute the relative trackline for mode $1$, the offset-follow behavior for mode $2$, return and surface for mode $3$, and the abort behavior for mode $4$.  Mission parameters for each vehicle for each of these modes are listed in table~\ref{tbl:mission6}.

\begin{figure}[!t] 
\centering
\includegraphics[width=1\textwidth]{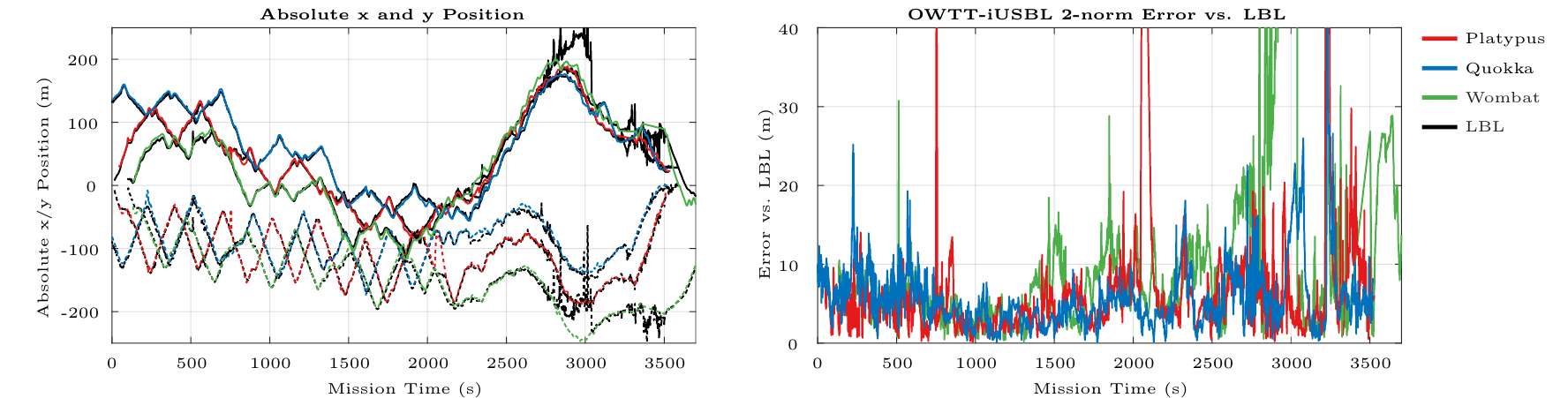}
\caption{Mission 6 plots of $x$ (solid) and $y$ (dashed) positions from OWTT-iUSBL navigation in color, and LBL localization in black (left); $2$-norm error of OWTT-iUSBL against LBL ground-truth (right).} 
\label{fig:mission6_error}
\end{figure}

Trajectories of the fleet during this mission are plotted in figure~\ref{fig:multi_experiment6}.  As with all six deployments during these three days, the boat and beacon were first driven out to a position in the river (in this case at about $(110,-70)$~\si{\meter}), and the beacon set to broadcast the mode $1$ signal; each AUV was then deployed from the dock by the third operator.  The middle plot of figure~\ref{fig:multi_experiment6} shows that each of the three vehicles correctly identified this mode, and subsequently entered the corresponding relative trackline behavior, carrying out these $120$~\si{\meter} long finite line transects at a heading of $160^\circ$, with a spacing of $35$~\si{\meter} between \emph{Platypus} and \emph{Quokka}, and $25$~\si{\meter} between \emph{Platypus} and \emph{Wombat}.  These transects were repeated continuously by the fleet while the motorboat re-positioned the beacon at two additional locations to the West -- the first at about $(35,-100)$~\si{\meter}, and the second at around $(-40,-125)$~\si{\meter}.  Re-positioning the beacon in this way enabled control over a multi-AUV survey of a large area of about $120$~\si{\meter} $\times$ $220$~\si{\meter}, in an approach very similar to the concept we illustrated in figure~\ref{fig:fleet_control} in section~\ref{sec:control}.  This relative trackline survey can be clearly seen in the absolute and relative position plots at the top-left of figure~\ref{fig:multi_experiment6}.

\begin{table}[b]
\caption{Modes and associated autonomous AUV behaviors and parameters for Mission 6.} \label{tbl:mission6}
\begin{center}
\begin{tabular}{|m{3cm}|c|c|c|c|}
  \hline
% after \\: \hline or \cline{col1-col2} \cline{col3-col4} ...
   \multicolumn{2}{|c|}{} & Platypus & Quokka & Wombat \\
  \hline\hline
  \multirow{4}{3cm}{Mode 1: \\Relative Trackline} & $(\Delta x, \Delta y)$ (m) & $(-14.1,-5.1)$ & $(18.8,6.8)$ & $(-37.6,-13.7)$ \\ \cline{2-5}
   & Line heading ($^\circ$) & $160.0$ & $160.0$ & $160.0$ \\ \cline{2-5}
   & Line length (m) & $120$ & $120$ & $120$ \\ \cline{2-5}
   & Buffer distance (m) & $14$ & $14$ & $14$ \\ \hline
  \multirow{3}{3cm}{Mode 2: \\Offset Follow} & $(\Delta x, \Delta y)$ (m) & $(7.5,-26.0)$ & $(-7.5,26.0)$ & $(22.5,-78.0)$ \\ \cline{2-5}
   & Buffer radius (m) & $15$ & $15$ & $15$ \\ \cline{2-5}
   & Depth ceiling (m) & $1$ & $1$ & $1$ \\ \hline
  \multirow{3}{3cm}{Mode 3: \\Return \& Surface} & $(\Delta x, \Delta y)$ (m) & $(0.0,-5.0)$ & $(2.2,-2.2)$ & $(-2.2,2.2)$ \\ \cline{2-5}
   & Line length (m) & $150$ & $150$ & $150$ \\ \cline{2-5}
   & Return heading ($^\circ$) & $340$ & $300$ & $20$ \\
  \hline
\end{tabular}
\end{center}
\end{table}

Around $1970$~\si{\second} into the mission, the operator switched the command to mode $2$, causing the fleet to switch to the offset follow behavior, with the AUVs attempting to maintain a line formation by positioning themselves at points $54$~\si{\meter} apart.  From its Western-most position, the boat slowly drove back to the East, then looped around to trace out a large `P'-shaped trajectory; the top-middle plots of figure~\ref{fig:multi_experiment6} clearly illustrate the three AUVs maintaining their beacon-relative positions during this period (as seen in the relative position plot), and consequently replicating the beacon trajectory with their sprint-and-drift behavior (occasionally circling back into position).  As a result, this line formation is fairly well maintained throughout this maneuver, as shown in the top-middle absolute position plot.  Note that the apex of this `P' trajectory occurs at the range limits of the LBL system, resulting in erroneous LBL solutions, outliers and drop-outs, most clearly shown in the bottom-right trajectory plot for \emph{Wombat}.  

Finally, as in the first mission, at about $3200$~\si{\second} the boat beacon was switched off, and the East LBL beacon was set to broadcast mode $3$, causing all three vehicles to return and surface at the dock.  Note the oscillatory behavior most visible in \emph{Wombat} during this return -- this is because that, unlike the relative trackline behavior, the return behavior does not have a buffer region to prevent such oscillations; additionally, although we calibrate the acoustic receiver to diminish angular biases due to local acoustic interactions, biases still remain and are most prominent at angles directly in front and behind the AUV, affecting navigation when the vehicle drives head-on towards the beacon.  Again, the individual vehicle plots at the bottom of figure~\ref{fig:multi_experiment6} demonstrate good qualitative agreement between OWTT-iUSBL navigation and the LBL solution.

\begin{figure}[!t] 
	\centering
	\includegraphics[width=0.75\textwidth]{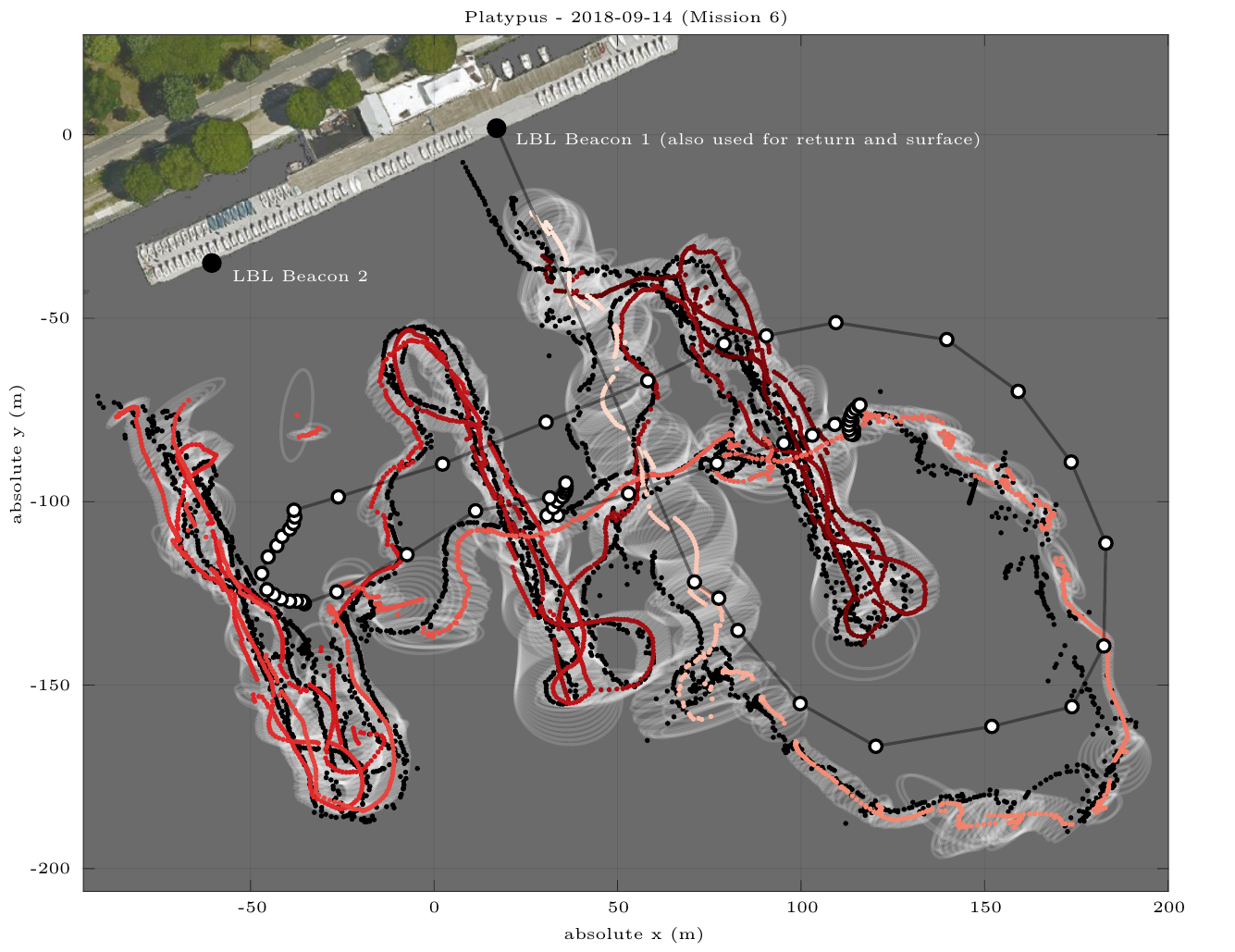}
	\caption{\emph{Platypus} trajectory from Mission 6 as estimated by OWTT-iUSBL navigation (red) and the LBL system (black) -- the OWTT-iUSBL trajectory is transformed into the absolute reference frame by offsetting \emph{Platypus'} on-board relative position estimate by the GPS position of the beacon in post-processing; beacon GPS position is shown as connected white dots, and $1\sigma$ covariance ellipses fitted to the particles are shown in semi-transparent white; only converged OWTT-iUSBL solutions are visualized, corresponding to a $1\sigma$ of $15$~\si{\meter} or less in both axes.} 
\label{fig:platypus_covariance}
\end{figure}

Plots of the $(x,y)$ position for each AUV are shown on the left of figure~\ref{fig:mission6_error}; it is again apparent that the OWTT-iUSBL and LBL solutions track each other very well.  Note that this plot clearly shows the fleet maintaining their coordinated line formation between $2200$~\si{\second} and $3200$~\si{\second} (while in mode $2$), with the AUVs tracking the desired $y$ separation of $52$~\si{\meter}.  LBL outliers that occur at the peak of the `P'-shaped track are visible as noisy LBL solutions between $2800$-$3050$~\si{\second} as well as $3200$-$3300$~\si{\second}, resulting in large $2$-norm errors at these times as shown in the right plot of figure~\ref{fig:mission6_error}.  Similarly to the first mission, the position error referenced against LBL for all three AUVs typically fluctuates around $10$~\si{\meter} or less; other error spikes occur due to the particle filter temporarily converging on an incorrect range mode (\emph{Platypus} at $750$~\si{\second}), other LBL outliers (\emph{Wombat} at $510$~\si{\second} and $1850$~\si{\second}), and unintended GPS fixes causing filter re-initialization (\emph{Platypus} at $2050$-$2100$~\si{\second}).  In contrast to OWTT-iUSBL navigation, the internal non-acoustically-aided dead-reckoning error grows unbounded, to a maximum of $45$~\si{\meter}, $120$~\si{\meter} and $210$~\si{\meter} for \emph{Platypus}, \emph{Quokka} and \emph{Wombat} respectively, illustrating the poor navigational performance of DVL-less AUVs in the absence of external aiding.

\emph{Platypus'} trajectory during mission $6$ is illustrated in more detail in figure~\ref{fig:platypus_covariance}, showing the OWTT-iUSBL position in red, as well as $1\sigma$ covariance ellipses from multivariate Gaussian fits to the particles in semi-transparent white; the ground-truth position from LBL is shown in black, while the GPS position of the beacon and boat are shown as connected white dots.  In terms of navigational accuracy, this plot illustrates that correct convergence of the filter results in good accuracy from OWTT-iUSBL localization as referenced against the LBL system, with the LBL solution for the most part falling within the $1\sigma$ uncertainty ellipse of the OWTT-iUSBL solution.  In terms of command and control, this figure illustrates that our relative navigation paradigm performs capably, with the operator reliably able to command \emph{Platypus'} behavior through mode selection, and to control the absolute position of planned trajectories through movement of the beacon.  The results presented here from mission $1$ and mission $6$ are representative of the performance of our OWTT-iUSBL system, and demonstrate the utility and power of our approach as a means of unifying localization, command, control and coordination under a single system to easily manage multi-AUV deployments with a small number of operators. 

\subsubsection{OWTT-iUSBL localization error statistics}

\begin{figure}[!t]
    \centering
    \includegraphics[width=1\textwidth]{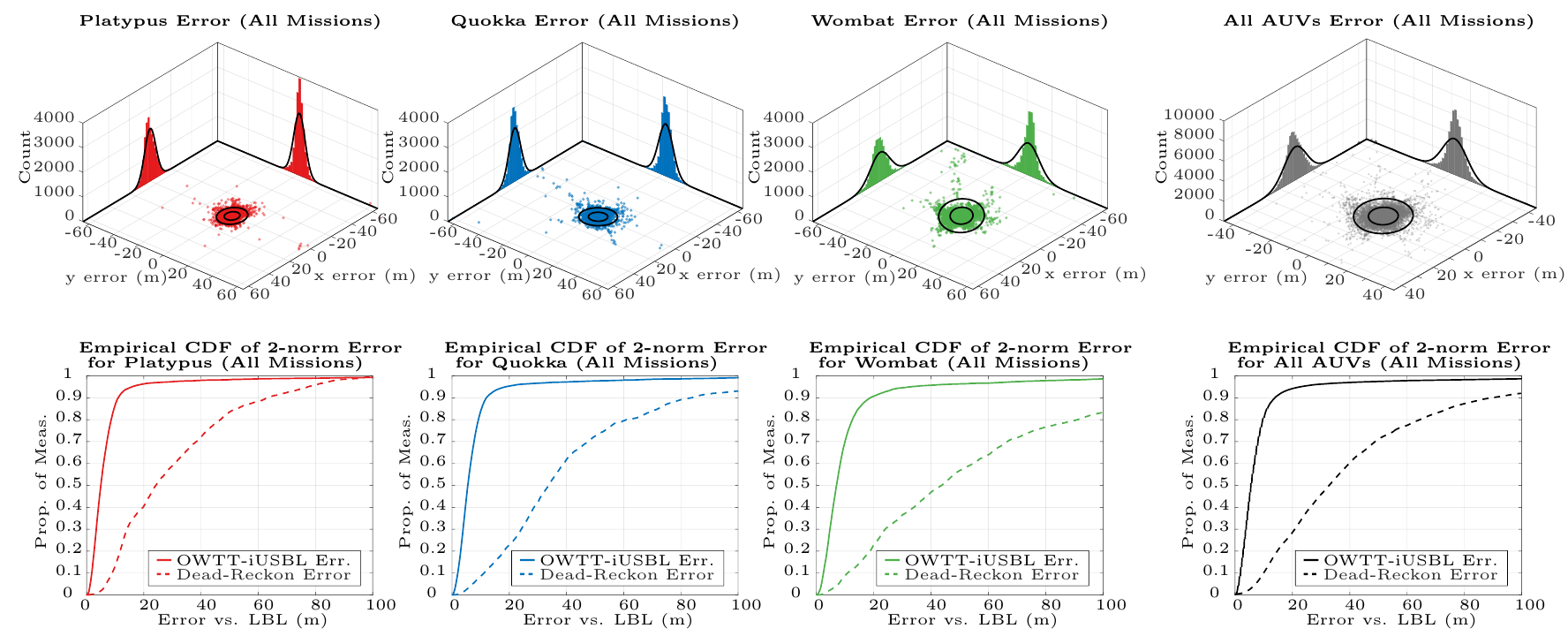}
    \caption{OWTT-iUSBL navigation and non-acoustically-aided dead-reckoning performance referenced against LBL ground-truth over all six missions -- the top row shows OWTT-iUSBL position error scatter plots and marginal $x$ and $y$ histogram distributions for individual vehicles and all vehicles combined, along with multivariate and univariate Gaussian fits; the bottom row shows corresponding empirical CDFs for the $2$-norm of error for both OWTT-iUSBL navigation (solid lines) and dead-reckoning (dashed lines).}
    \label{fig:localization_stats}
\end{figure}

%%%TABLE%%%
\newcolumntype{C}{>{\centering\arraybackslash}m{3em}}
\begin{table}[b]
\caption{Statistics of OWTT-iUSBL navigation error referenced against LBL for all six multi-AUV missions.} \label{tbl:position_stats}
  \begin{center}
  \begin{tabular}{ | r | c | c | c | c | c || c | c |  }
  \hline
  \multicolumn{1}{|c|}{} & \multicolumn{5}{c||}{$(x,y)$ Error} & \multicolumn{2}{c|}{$2$-norm Error} \\
  \hline
  \multicolumn{1}{|c|}{} & $\mu_{x,y}$ & $\sigma_{major}$ & $\sigma_{minor}$ & $\sigma_{x}$ & $\sigma_{y}$ & $68\%$-tile & $95\%$-tile \\ 
  \hline
  \hline
  \rule{0pt}{2ex} Platypus & (0.205, 1.551) & 5.433 & 4.604 & 5.425 & 4.613 & 6.85 & 15.81 \\ 
  \hline
  \rule{0pt}{2ex} Quokka & (0.539, 2.754) & 6.001 & 5.526 & 5.689 & 5.846 & 7.45 & 18.52 \\ 
  \hline 
  \rule{0pt}{2ex} Wombat & (0.113, 1.223) & 10.430 & 7.080 & 9.368 & 8.436 & 9.66 & 32.75 \\ 
  \hline
  \hline
  \rule{0pt}{2ex} All AUVs & (0.294, 1.874) & 7.375 & 6.032 & 6.987 & 6.477 & 7.89 & 22.74 \\ 
  \hline
  \end{tabular}
  \end{center}
\end{table}
%%%TABLE%%%

To quantify the accuracy of the OWTT-iUSBL system for localization, error statistics of the system as referenced against LBL ground-truth were gathered over all six deployments that took place on the 10\textsuperscript{th}, 12\textsuperscript{th} and 14\textsuperscript{th} of September 2018.  These statistics are plotted in figure~\ref{fig:localization_stats}, with scatter plots of these errors shown at the top of this figure for \emph{Platypus}, \emph{Quokka} and \emph{Wombat}, as well as from all AUVs combined, along with $1\sigma$ and $2\sigma$ covariance ellipses from Gaussian fitting; marginal error distributions over $x$ and $y$ are shown as histograms on the `walls' of these plots, also with Guassian fits.  The bottom row of figure~\ref{fig:localization_stats} show the corresponding empirical CDFs of the $2$-norm error for each vehicle and for all AUVs combined -- CDFs are shown for both OWTT-iUSBL navigation (solid lines) and for the internal non-acoustically-aided dead-reckoning estimate (dashed lines).

Major statistical values from these plots are listed in table~\ref{tbl:position_stats}; the mean error in $x$ and $y$ for each AUV and all AUVs combined are recorded in the first column, as well as the $1\sigma$ standard deviations along the major ($\sigma_{major}$) and minor ($\sigma_{minor}$) axes of the multivariate Gaussian fits to the $(x,y)$ data, and the $1\sigma$ standard deviation of the Guassian fits to the marginal $x$ and $y$ distributions ($\sigma_{x}$ and $\sigma_{y}$ respectively).  Interestingly, we can see that \emph{Platypus'} navigational performance is the superior of the three vehicles, while \emph{Wombat} is the worst performer; this trend is confirmed by the $2$-norm CDF error statistics listed in the last two columns of table~\ref{tbl:position_stats}.  $68\%$ of \emph{Platypus'} OWTT-iUSBL position errors fall below $6.85$~\si{\meter}, while for \emph{Quokka} and \emph{Wombat}, this figure is $7.45$~\si{\meter} and $9.66$~\si{\meter} respectively; this relative performance between vehicles can be confirmed by looking at the CDF curves in the lower plots of figure~\ref{fig:localization_stats}.  However, note that this trend is also reflected in the dead-reckoning CDF curves from these plots -- this suggests that dead-reckoning and OWTT-iUSBL performance are interrelated; we surmise that this is due to limited accuracy in the AUV heading estimate, since inaccurate heading has a significant impact on both OWTT-iUSBL and dead-reckoning localization.

Across all AUVs, the statistics listed in table~\ref{tbl:position_stats} indicate that OWTT-iUSBL positioning under the relative acoustic navigation operating paradigm is fairly accurate -- $68\%$ of errors in position fall under $7.89$~\si{\meter}, with $x$ and $y$ standard deviations of $6.99$~\si{\meter} and $6.48$~\si{\meter} respectively.  It should be noted that a number of sources contribute to these error statistics, and so errors should not be attributed solely to the OWTT-iUSBL system -- GPS error in beacon position, as well as error in the LBL system both contribute.  Nevertheless, these statistics, combined with previously presented trajectory plots, provide convincing support for the fact that the OWTT-iUSBL system is able to provide accurate localization for multi-AUV deployments.

\section{Discussion and conclusion}\label{sec:conclusion}

In this article, we have described a unified approach to localization, command, control and coordination for multiple low-cost autonomous underwater vehicles (AUVs), which lack conventional navigational sensors such as the Doppler velocity log (DVL) due to constraints in platform size and available energy.  Results from six deployments of a fleet of three Bluefin SandShark AUVs over three separate days within a shallow-water river environment in September 2018 were presented, providing compelling evidence of the capability of this approach for managing multi-AUV deployments, and for providing accurate navigation for multiple AUVs in a scalable manner.  This approach is centered around the use of a custom-built one-way travel-time inverted ultra-short baseline (OWTT-iUSBL) positioning system, with OWTT ranging from clock-synchronization and angle estimation from time-delays between elements of the USBL receiver enabling each AUV to determine distance and angle to a single, periodically-broadcasting navigation beacon.  The OWTT-iUSBL system makes use of these acoustic range and angle measurements, derived from matched filtering and beamforming, within a particle filter framework running on each AUV for online navigation.  

Comparing online position solutions from each AUV to those of a secondary long baseline (LBL) system in post-processing demonstrated that $68\%$ of the difference in solutions fell below $7.9$~\si{\meter}, with the standard deviation of these differences in $x$ and $y$ being approximately $7.0$~\si{\meter} and $6.5$~\si{\meter} respectively.  These results indicate that the OWTT-iUSBL system is able to provide good accuracy localization with bounded error, especially considering the acoustically-challenging environment of the river.  Since the USBL receiver on each vehicle passively records and does not transmit acoustically, the system is low-power and can provide a navigation solution to any number of AUVs within receive range of the acoustic beacon.

%With each vehicle having an understanding of their position relative to the beacon, fleet-wide control of absolute position is made possible by designing AUV behaviors within the beacon-centric reference frame, and by re-positioning the beacon within the absolute frame of reference.  Coordination between vehicles is achieved by specifying parameters of these behaviors in the context of the fleet to avoid inter-vehicle collisions, while different AUV behaviors are commanded by having the beacon transmit different signals from within a set.

Fleet-wide command and control was successfully demonstrated over six separate deployments in the Charles River -- a two-person team operating a motorboat with the beacon were able to command the three-vehicle SandShark AUV fleet to perform different behaviors by manually switching between various broadcast linear frequency-modulated chirps.  Since each AUV is limited to only estimating the position of the beacon relative to themselves, a number of vehicle behaviors were designed within a beacon-centric frame of reference; these behaviors included a relative loiter, offset follow, relative trackline, and return and surface behavior, with prototypes of these behaviors each successfully performed over the six experimental deployments.  Control over absolute position of the fleet was demonstrated in these experiments by movement of the beacon by the two-person motorboat team, causing vehicles in the fleet to follow the beacon in order to maintain the relative position of the commanded behavior.

Control over fleet-wide position enabled the fleet to perform coordinated behaviors, such as an area survey and a moving line formation.  This coordination was achieved without inter-vehicle communication, by uniquely specifying the parameters of our custom behaviors on each AUV within the context of the fleet -- for example, setting different values for the radius of the relative loiter, or offsets for the relative trackline and offset follow behaviors to prevent vehicle collisions.  By doing so, we demonstrated that the relative trackline behavior could be used to survey an area of about $120$~\si{\meter} $\times$ $220$~\si{\meter} with the fleet, by moving the beacon perpendicular to the direction of the tracklines;  later on in the same mission, the fleet was commanded into the offset follow behavior, which was used to demonstrate coordinated formation keeping, with the vehicles maintaining a line while following a `P'-shaped trajectory.

It is interesting to note some of the advantages and disadvantages of the localization approach used here with prior OWTT positioning methods -- one drawback of OWTT-iUSBL localization as described here is the added necessity of acoustic calibration of the USBL receivers, which is not required by prior approaches such as single-beacon OWTT ranging and OWTT long-baseline (LBL) positioning.  As a trade-off, OWTT-iUSBL enables single-beacon localization with a fully-determined position estimate on every transmission, in exchange for this calibration process and the algorithmic complexity required for angle estimation.  On the other hand, single-beacon OWTT ranging provides a localization method that is simpler in terms of hardware (since an array is unnecessary), but requires multiple range measurements for an unambiguous position fix, ruling out its use with our relative navigation approach.  Finally, OWTT LBL also provides a full position solution every cycle, but complicates the deployment process with the need for multiple beacons, and requires the transmission of beacon position information if the baseline is moving.

The experimental results presented here are intended to demonstrate that our OWTT-iUSBL relative navigation approach can feasibly and capably manage multi-AUV deployments in a convenient manner; and consistent performance over repeated deployments have illustrated that our approach is robust both operationally and in terms of localization performance.  However, future work can provide significant improvements to the system: currently, a particle filter provides online AUV navigation -- rather than use the particle filter estimate for absolute geolocalization, the use of batch optimization techniques such as nonlinear least-squares would improve the accuracy of vehicle trajectories in post-processing;  the prototype AUV behaviors presented here were intended to provide a simple demonstration of command and control of the fleet -- their performance can be improved through the use of sophisticated control techniques to account for uncertainty in the relative beacon position, and to synchronize fleet movement through time-parameterization of planned paths; finally, the AUVs in the fleet do not communicate, and are unable to provide operators with information about their status -- the integration of acoustic modems, used sparingly to prevent saturation of the acoustic channel, could improve fleet coordination and allow operators to effectively monitor fleet status.

As low-cost AUV platforms become more readily available, deployments consisting of fleets of tens of vehicles or more are becoming a reality.  As the number of AUVs in the fleet grows, the conventional operating paradigm for underwater vehicles becomes impractical and increasingly infeasible without a sizeable team of personnel to manage the fleet -- individual vehicle trajectories must be pre-planned and programmed, and physical limitations on the acoustic channel mean that each AUV has an increasingly smaller amount of time or bandwidth in the channel for receiving operator commands, to relay coordination demands between themselves, and to transmit status information to their operators.  The OWTT-iUSBL relative navigation operating paradigm presented in this work has shown significant potential in providing an effective approach for managing the deployment of multiple AUVs in a manner that is convenient, that requires few operators, and that unifies localization, command, control and coordination within a single system.  Near-term future work will focus on expanding our fleet of vehicles and on sensor integration, with the goal of using this fleet and our approach to perform synoptic measurement of submesoscale ocean features that exhibit spatiotemporal aliasing such as chemical plumes, internal waves, or temperature fronts.

% localization smoothing offline, behavior improvements (prototype, conceptual, add time parameterization), acoustic modems for status, larger fleet, use for sampling of submesoscale ocean phenomena starting with conductivity and temperature fronts.

% as multi-AUV becomes a reality, operating larger numbers becomes impractical/infeasible for small number of operators, becomes costly. 

\subsubsection*{Acknowledgments}
The authors would like to sincerely thank the many members of the LAMSS group at MIT who assisted in AUV deployments and system testing, including Dr.~M.~Benjamin, Dr.~M.~Novitzky, Dr.~P.~Robinette, O.~Viquez, Dr.~S.~A.~T.~Randeni~P., G.~Nannig, R.~Chen, and E.~Bhatt, as well as the MIT Sailing Pavilion staff for use of their facilities.   We also thank the anonymous reviewers for their critical comments.  This work was partially supported by the Office of Naval Research, the Defense Advanced Research Projects Agency, Lincoln Laboratory, and the Reuben~F. and Elizabeth~B.~Richards Endowed Funds at WHOI.

\bibliographystyle{apalike}
\bibliography{multiAUV_rypkema_refs}

\end{document}